\documentclass{article}

\usepackage[utf8]{inputenc}
\usepackage[T1]{fontenc}
\usepackage[english]{babel}
\usepackage{arxiv}
\usepackage{amsmath, amssymb, amsfonts, amsthm}
\usepackage{bm}
\usepackage{microtype}
\usepackage{graphicx}
\usepackage{booktabs}
\usepackage{multirow}
\usepackage{makecell}
\usepackage{threeparttable}
\usepackage{tabularx}
\usepackage{xcolor}
\usepackage[colorlinks=true, allcolors=blue]{hyperref}
\usepackage{url}
\usepackage{doi}
\usepackage{natbib}
\bibpunct{[}{]}{;}{a}{,}{,}
\usepackage{cleveref}

\newtheorem{proposition}{Proposition}

\theoremstyle{remark}

\newcommand{\E}{\mathbb{E}}

\newcommand{\bpb}{\mathrm{bpb}}
\newcommand{\MASK}{\texttt{[MASK]}}
\newcommand{\EOS}{\texttt{[EOS]}}

\title{Homebrewed SLMs: Autoregressive and Masked Diffusion Language Models Built at Home for Portuguese}

\author{\href{https://orcid.org/0000-0002-1288-7695}{\includegraphics[scale=0.06]{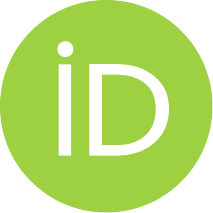}\hspace{1mm}Pedro Carvalho Brom} \\
	L-PRISMA, Department of Mathematics\\
	Federal Institute of Brasília\\
	Brasília, Brazil \\
	\texttt{pedro.brom@ifb.edu.br} \And
\href{https://orcid.org/0000-0003-1826-1850}{\includegraphics[scale=0.06]{orcid.pdf}\hspace{1mm}Li Weigang} \\
	TransLab, Department of Computer Science\\
	University of Brasília\\
	Brasília, Brazil \\
	\texttt{weigang@unb.br}
}

\renewcommand{\shorttitle}{Homebrewed SLMs}
\hypersetup{
pdftitle={Homebrewed SLMs: Autoregressive and Masked Diffusion Language Models Built at Home for Portuguese},
pdfsubject={cs.CL},
pdfauthor={Pedro Carvalho Brom, Li Weigang},
pdfkeywords={homebrewed language models, small language models, masked diffusion, Portuguese, tutorial, reproducibility, evaluation, specialist domain},
}

\usepackage[section]{placeins}

\let\oldsubsection\subsection
\renewcommand{\subsection}{\FloatBarrier\oldsubsection}

\begin{document}
\maketitle

\begin{abstract}
Building a language model of one's own no longer requires a data centre. This tutorial trains one from scratch, tokenizer included, on one consumer GPU, and shows how to tell whether it is good for the tasks of a specialist domain. It proposes no new method. It releases a tested Python package and two notebooks in Brazilian Portuguese with autoregressive and masked diffusion generation on one network, a preregistered comparison protocol, legal tasks built from text alone with a Kneser--Ney floor and public yardsticks, reference numbers for models of 11 to 12M parameters, and a catalogue of defects that only measurement revealed. On Brazilian prose and federal laws, at minutes per run, the autoregressive mode peaks at 64 epochs over the same text and then memorises it, while masked diffusion improves up to 256. At their best points on test, the diffusion bound exceeds the autoregressive likelihood by 0.06 bits per byte, within the preregistered minimum effect. On held-out laws and on acts published after training, the model ties a 6-gram floor in likelihood on the same windows, and it loses to the floor at continuing those acts. It beats the floor at detecting tampered text, and fills cross-references more often than the floor and public models of 160M and 630M parameters, only in decrees, a genre absent from training, where those scorers stay near or below chance, and in filling the blank only when each group weighs the same, since per item it ties the 630M model. On laws, the public models, trained with about $5 \times 10^2$ to $10^5$ times its compute, lead in tampered text and continuation. The blind human reading is preregistered and not part of this version. Code and notebooks: \url{https://github.com/pcbrom/homebrewed-slms}.
\end{abstract}


\section{Introduction}\label{sec:intro}

Many people who need a language model for a narrow domain, such as the laws of one country, the records of one hospital or the literature of one period, face the same question. Is it possible to train a model of one's own, from nothing, on a home computer, and obtain something that is actually good for the tasks at hand? Most learning material does not answer it, because it starts from a model someone else pretrained. A checkpoint is downloaded and a library fine-tunes it, and the learner rarely sees where the loss comes from, how the tokenizer was built, or how to know whether a fluent sample reflects learning or recitation of the training text. Adapting a small pretrained model remains a strong road when one exists for the language. \citet{di2024slim} fine-tuned TeenyTinyLlama \citep{correa2024teenytiny}, a small model pretrained on Brazilian Portuguese, to the Mercosur Common Nomenclature, a specialist customs domain, and compared it with GPT-4 on questions held out from training. This tutorial takes the other road, training from nothing, which is the one left when no pretrained model fits the language, the period or the vocabulary of the domain, and the one that keeps every step visible.

This tutorial answers the question with a complete, measured example. It builds a small language model (SLM) by hand, from the raw text to the generated samples, and evaluates it against references that give meaning to its numbers. The reader who follows it ends with three things. The first is a working pipeline, a Python package and two notebooks that run each step with its input and output shown. The second is a way to adapt the pipeline to a corpus of their own by editing a configuration and a domain adapter. The third is an evaluation protocol that says whether the resulting model is good, compared with what, and for which tasks.

The tutorial proposes no new method, and its value lies in putting known pieces together where each can be checked. Masked diffusion in the number-of-masks form follows \citet{ou2024radd} and \citet{nie2025llada}, the floor is interpolated modified Kneser--Ney smoothing \citep{chen1996smoothing}, the statistics are standard, and the protocol borrows preregistration and checks before results from good practice in empirical research \citep{vanmiltenburg2021prereg,dodge2019show}. What a reader can reuse, and cite, is the assembly, in five parts.
\begin{description}
\item[C1, package and notebooks.] The \texttt{slm} package, eleven modules each guarded by tests (\Cref{tab:boxes}), and two notebooks that run it step by step, one per generation mode, written for the free GPU of Colab. The reproduction on that GPU is not part of this version. Code identifiers and notebooks are in Portuguese, and \Cref{tab:boxes} glosses each module.
\item[C2, comparison protocol.] Checks with criteria fixed before the measurement, preregistrations frozen by their SHA-256 hash, a minimum effect, a seed count derived from a pilot and dated amendments (\Cref{sec:protocol}, folder \texttt{preregistro}).
\item[C3, task kit.] Four legal tasks built by a script from the text of 1,094 federal acts published in 2025 and 2026, with no human labelling, the item hashes and the scoring rule, which place a home-built model between a Kneser--Ney floor and public models (\Cref{sec:tasks}, folder \texttt{tarefas}).
\item[C4, reference numbers.] Bits per byte, task scores and calibration for models of 11 to 12M parameters trained on a few megabytes of text, with seeds, intervals and the floor scored on the same windows, each reproducible on one GPU in minutes (\Cref{sec:results}, folders \texttt{runs} and \texttt{resultados}).
\item[C5, catalogue of defects.] The defects that only measurement revealed, each with the check that detected it and the test that now guards it (\Cref{sec:checks}).
\end{description}

The model has two generation modes on one network. The autoregressive mode predicts each token from the previous ones with causal attention. The masked diffusion mode \citep{austin2021d3pm,sahoo2024mdlm,shi2024md4,ou2024radd,nie2025llada} predicts masked tokens from the visible ones with bidirectional attention. The only differences between the two are the attention mask and the loss, so the tutorial can teach both with the same code, and a measured difference between them comes from training and generation, not from the network.

The worked example uses two Portuguese corpora. The first, the prose of Machado de Assis, has a recognisable style and serves to demonstrate each step. The second, Brazilian federal legislation, is the specialist domain on which usefulness is measured. All runs used one consumer GPU under a 500\,W power cap, one run at a time.

The tutorial is organised in the order in which a reader would work. \Cref{sec:related} places it among comparable resources. \Cref{sec:needs} states what is needed in hardware, data and time. \Cref{sec:background} gives the two training objectives. \Cref{sec:pipeline} builds the model step by step, with a note on adapting each step to another domain. \Cref{sec:evaluation} explains how to evaluate the model, which is the centre of the tutorial. \Cref{sec:protocol} explains how to compare configurations so that a difference can be trusted. \Cref{sec:results} reports the worked example. \Cref{sec:checks} lists the checks that kept defects out of the numbers, and \Cref{sec:adapting} gathers the steps to move the pipeline to a new domain.

\section{Related resources}\label{sec:related}

Several resources already teach how to build a language model by hand, and a reader choosing among them needs to know what each one covers. \Cref{tab:related} compares them on axes that can be checked in their papers and repositories. The Annotated Transformer \citep{rush2018annotated} set the format of code that reads alongside the paper it implements. nanoGPT \citep{karpathy2022nanogpt} is the reference for a short autoregressive training loop. Cramming \citep{geiping2022cramming} trains a masked encoder on one GPU in one day, with a tokenizer built from its own data. TinyStories \citep{eldan2023tinystories} shows that models under 10M parameters can produce fluent text on a simplified synthetic corpus, graded by a large model, and the BabyLM challenges \citep{warstadt2025babylm,hu2024babylm2} compare pretraining on about 10M to 100M words under a shared evaluation, where in the second edition a hybrid causal and masked model ranked first. None of these covers a language other than English, two generation modes on one network, a floor and tasks for a specialist domain, or preregistration. Those four are what this tutorial adds, and in return it covers less ground than any of them on scale, on architecture variants and on breadth of benchmarks.

\begin{table}[!htbp]
\centering
\caption{Comparable resources on axes that can be checked in their papers and repositories.}
\label{tab:related}
\scriptsize
\begin{tabularx}{\linewidth}{@{}>{\raggedright\arraybackslash}p{2.2cm}>{\raggedright\arraybackslash}p{1.5cm}>{\raggedright\arraybackslash}X>{\raggedright\arraybackslash}X>{\raggedright\arraybackslash}X>{\raggedright\arraybackslash}p{1.2cm}>{\raggedright\arraybackslash}p{1.6cm}@{}}
\toprule
Resource & Language & Tokenizer & Generation & Evaluation & Prereg. & Code licence \\
\midrule
Annotated Transformer & English, German & BPE of the original setup & Encoder--decoder & BLEU & No & MIT \\
nanoGPT & English & GPT-2 BPE reused, or characters & Autoregressive & Validation loss & No & MIT \\
Cramming & English & WordPiece trained on the data & None, masked encoder & GLUE & No & MIT \\
TinyStories & English, synthetic & GPT-Neo BPE reused & Autoregressive & Grading by GPT-4 & No & Data CDLA-Sharing 1.0 \\
BabyLM & English & Participant's choice & Participant's choice & BLiMP, (Super)GLUE and others & No & Varies \\
This tutorial & Portuguese & Byte-level BPE trained on the data & Autoregressive and masked diffusion, one network & Kneser--Ney floor, public yardsticks, four domain tasks, calibration & Yes, hashed & CC BY-NC-SA 4.0 \\
\bottomrule
\end{tabularx}
\end{table}

\section{What is needed}\label{sec:needs}

\paragraph{Hardware.} The reference machine has one NVIDIA RTX 5090 with a power cap of 500\,W. Training runs in fp16 with gradient scaling, which also runs on the free T4 GPU of Google Colab, where the notebooks are meant to be used in class. The reproduction of the reference runs on the T4, with its time per step, is not part of this version.

\paragraph{Software.} Python with PyTorch, the \texttt{tokenizers} library at a pinned version, NumPy and SciPy. Corpus construction uses only the standard library. Every run records the versions in use.

\paragraph{Data.} A corpus of a few megabytes of clean text in the domain of interest, divided into documents. The worked example uses 2.22\,MB of training text in the first corpus and 4.31\,MB in the second.

\paragraph{Time and energy.} On the reference machine, training processed 1.13M tokens per second in the autoregressive mode and 1.04M in diffusion. A run of 16 epochs over the first corpus took about 12 seconds of training and a run of 256 epochs about 3 minutes. The full study of the first corpus, 158 runs including the hyperparameter search, the pilot and the comparison, took 3.94 hours of training. The power cap times that time gives 1.97\,kWh, an upper bound for the GPU alone, which leaves out the rest of the machine. On corpus B, with a smaller vocabulary, a run at the best budget of each mode took about 90 seconds in the autoregressive mode at 64 epochs and 7 minutes in diffusion at 256 epochs, and the 56 runs of its budget grid took 1.66 hours.

\section{Background}\label{sec:background}

\subsection{Autoregressive likelihood}

For a sequence $x = (x^1, \dots, x^L)$ of tokens, the autoregressive mode factorises the probability by the chain rule and minimises the negative log-likelihood
\begin{equation}
-\log p_\theta(x) = -\sum_{i=1}^{L} \log p_\theta(x^i \mid x^{<i}),
\end{equation}
computed in one pass with causal attention and the input shifted by a \texttt{[BOS]} token, so that the prediction of $x^i$ sees only $x^{<i}$. The value is exact.

\subsection{Masked diffusion and its bound}

Masked diffusion corrupts a sequence by replacing tokens with an absorbing \MASK{} token and learns to undo the corruption \citep{austin2021d3pm}. With the linear schedule $\alpha_t = 1 - t$, each token is masked independently with probability $t \in (0, 1]$, and the negative evidence lower bound (NELBO) on $-\log p_\theta(x)$ takes the form
\begin{equation}\label{eq:nelbo-t}
\mathcal{L}(x) = \int_0^1 \frac{1}{t}\, \E_{x_t}\Big[\sum_{i:\, x_t^i = \MASK} -\log p_\theta(x^i \mid x_t)\Big]\, dt
\end{equation}
\citep{sahoo2024mdlm,shi2024md4}, a line that also includes score-based discrete diffusion \citep{lou2023sedd}. \citet[Theorem 1]{ou2024radd} showed that, for the absorbing process, the concrete score factorises into a scalar that depends only on time and the distribution of the clean tokens given the visible context, which does not depend on time. The network therefore needs no time input, and RADD and LLaDA \citep{ou2024radd,nie2025llada} use an ordinary Transformer without one. This is what allows the two modes of the tutorial to share the same network.

Conditioning \eqref{eq:nelbo-t} on the number $l$ of masked positions gives the form used throughout. Write $F_l = \E_{S}\big[\sum_{i \in S} -\log p_\theta(x^i \mid x_{\setminus S})\big]$ for the expected loss when a set $S$ of exactly $l$ positions, drawn uniformly, is masked.

\begin{proposition}[Number-of-masks form]\label{prop:nmasks}
$\mathcal{L}(x) = \sum_{l=1}^{L} F_l / l = \E_{l \sim U\{1,\dots,L\}}\big[(L/l)\, F_l\big]$.
\end{proposition}

The single-sample estimator masks exactly $l$ positions, with $l$ drawn uniformly and stratified in the batch, and weighs the summed cross-entropy by $L/l$. This is the form of \citet[Theorem 2 and Appendix C]{ou2024radd} and of \citet[Eq.~14]{nie2025llada}. It needs no lower limit on $t$, never draws the useless case $l = 0$, and its weight never exceeds $L$.

\begin{proposition}[Smaller second moment]\label{prop:variance}
Let $f_l$ be the summed cross-entropy over a uniformly drawn set of $l$ masked positions. The time-form estimator $(1/t)\, f_l$, with $t \sim U(0,1]$ and $l \mid t \sim \mathrm{Bin}(L, t)$, and the number-of-masks estimator $(L/l)\, f_l$, with $l \sim U\{1,\dots,L\}$, have the same mean $\mathcal{L}(x)$. The second moment of the time form is at least $(L+1) \sum_{l} \E[f_l^2]/l^2$, and that of the number-of-masks form equals $L \sum_l \E[f_l^2]/l^2$.
\end{proposition}

Both proofs are in \Cref{app:proofs}. \Cref{prop:variance} is a short consequence of \Cref{prop:nmasks} and a Rao--Blackwell step, stated here because it explains the choice of estimator, and is not claimed as a new method. The time form also has infinite variance without a cut-off $t \ge \varepsilon$, and the cut-off removes a mass of order $L\varepsilon$ from the bound. Training divides the per-sequence estimator by $L$, so the per-sequence loss is $(1/l)\sum_{i \in S} -\log p_\theta(x^i \mid x_{\setminus S})$, averaged over the batch. Normalising by the number of masked tokens is a different objective and is not used.

\section{Building the model step by step}\label{sec:pipeline}

The pipeline has eleven steps, called boxes and numbered 0 to 10 (\Cref{tab:boxes}). Each box is at once a module of the package and a section of each notebook, and each module depends only on the boxes before it. The notebooks import the package instead of reimplementing it, so the code that trained the reported runs is the code a student executes in class. Each notebook section has three parts, the theory with its formulas, the executed code with input and output, and the technique, with the problems met in the project and the measurement that revealed them.

\begin{table}[!htbp]
\centering
\caption{The eleven boxes of the pipeline, their modules with an English gloss of each Portuguese name, and the checks that guard them. The decision documents of the repository define each check.}
\label{tab:boxes}
\small
\begin{tabularx}{\linewidth}{@{}clXl@{}}
\toprule
Box & Module (gloss) & Responsibility & Checks \\
\midrule
0 & \texttt{config} (configuration) & Validated configuration of corpus, tokenizer, network, mode and budget & \\
1 & \texttt{dados} (data) & Domain adapters, cleaning, segmentation, block split, proper-name lists & Corpus G0 to G5 \\
2 & \texttt{tokenizador} (tokenizer) & Tokenizer contract, didactic BPE, library adapter & Tokenizer G0 to G5 \\
3 & \texttt{janelas} (windows) & Windows of $L$ tokens per document, with \EOS{} and padding predicted & Formulation G1, G2 \\
4 & \texttt{mascaramento} (masking) & Stratified draw of $l$ and masking of exactly $l$ positions & Formulation G0 \\
5 & \texttt{modelo} (model) & Transformer with causal or bidirectional attention, \MASK{} logit at $-\infty$ & Formulation G0 \\
6 & \makecell[l]{\texttt{perda} (loss),\\ \texttt{treino} (training)} & Losses, AdamW, fp16 with gradient scaling, budget per applied step, sentinel & Precision G1 to G4 \\
7 & \texttt{geracao} (generation) & Ancestral sampling with random unmasking, draws in float64 & Formulation G5 \\
8 & \texttt{avaliacao} (evaluation) & Bits per byte and NELBO with $K$ samples & Formulation G3 \\
9 & \texttt{estatistica} (statistics) & Two-level bootstrap, randomization test, TOST, seed count & Budget G3 to G6 \\
10 & \texttt{leitura} (reading) & Lexical level, copy filter, blind batches, agreement & Budget G7 \\
\bottomrule
\end{tabularx}
\end{table}

\subsection{Corpus}

The first corpus (A) is the prose of Machado de Assis in the Project Gutenberg, ten works after removing \emph{Poesias completas} and \emph{Relíquias de casa velha}, the latter because it contains a sonnet and two comedies. Cleaning keeps the original spelling and normalises the text to NFC once, before anything else. The split uses contiguous blocks of whole chapters per work, one for validation and one for test, separated by a discarded margin chapter, with at least 5 chapters and 30\,KB per block, so that neighbouring chapters never fall on both sides. Nine works enter the test partition, and \emph{Papéis avulsos} stays in training and validation only, because its twelve stories cannot hold two blocks of five, two margins and a training block. Training holds 2.22\,MB and 624,377 tokens, validation 153,756 tokens and test 133,520.

The specialist corpus (B) is Brazilian federal legislation, ordinary laws from 2019 to 2022 of the substantive and amending classes. Laws that open budget credits, set budgets or pay homage were removed, because their text repeats a formula from one law to the next. The corpus holds 4.76\,MB in 441 laws, collected from the official compiled pages with struck-out text and editorial notes removed. All 444 pages of the two classes were collected, and three laws left the corpus because they are wholly repealed and their compiled page has no text in force. Each law joins the group of the first law cited in its summary, near-copies are merged by MinHash, and a whole group goes to a single partition, so that a law and the law it amends do not fall on both sides. Both texts are in the public domain or are official acts without copyright protection under Brazilian law, and the repository versions them with a manifest of URLs, hashes and collection dates, so the notebooks download nothing beyond the repository. The largest group of corpus B, Law 14,133 on public procurement, holds 5.78\% of its bytes. A third set, built by the same adapter, holds the 300 ordinary laws of the same two classes published in 2025 and 2026, 163 amending and 137 substantive, 2.36\,MB in 229 groups, where the largest group holds 10.9\% of the bytes. No model in this tutorial can have seen these laws, neither the one trained here, whose corpus ends in 2022, nor the public models used as yardstick, all published before 2025, so this external set carries the evaluation of usefulness. The tasks of \Cref{sec:tasks} enlarge it with the complementary laws and decrees of the same years.

\emph{Adapting to another domain.} Write an adapter that turns the raw sources into documents with an identifier, a group and the cleaned text. The group is the unit that must never be split across partitions, such as a work, a law and the laws it amends, or a patient. Measure the partitions before training anything.

\subsection{Tokenizer}

The tokenizer is byte-level BPE \citep{sennrich2016bpe} with the GPT-2 pre-tokenisation pattern \citep{radford2019gpt2}, trained only on the training partition. A tokenizer from a library is fast, and one written by hand teaches the algorithm, so the tutorial keeps both and requires them to agree. With a written tie-breaking convention, the didactic BPE and the \texttt{tokenizers} library 0.23.3 produce the same merges and the same ids on held-out text. The vocabulary size is chosen before any model training as the largest value in $\{1024, 2048, 4096, 8192\}$ with at most 10\% of the learned tokens absent from the held-out text, which keeps rarely seen rows out of the embedding \citep{land2024magikarp}. The rule gives $V = 4096$ on corpus A and $V = 1024$ on corpus B. The tokenizer sits behind a contract of six tests, and a trivial tokenizer of 260 byte ids passes them without any change to the package, so another tokenizer can replace BPE.

\emph{Adapting to another domain.} Train the tokenizer on the new training partition and apply the vocabulary rule to its own held-out text. A smaller or more repetitive corpus will usually select a smaller vocabulary, as corpus B did.

\subsection{Windows}

Each document is cut into windows of exactly $L = 256$ tokens, aligned to its start, with \EOS{} at the end and padding up to $L$. Both modes predict every position, padding included, with the nats counted and zero bytes, and documents are never concatenated. Padding occupies 12.2\% of the training positions at $L = 256$, against 6.3\% at 128 and 22.5\% at 512, a measurement made before fixing $L$.

\subsection{Network}

A pre-LayerNorm Transformer with 6 layers, $d = 384$, 6 heads, learned positions, a GELU MLP of width $4d$, embedding tied to the output layer \citep{press2016tying} and no time input. It has 10.6M non-embedding parameters, 12.3M in total with $V = 4096$ and 11.1M with $V = 1024$. The \MASK{} logit is fixed at $-\infty$ in both modes, so the model never predicts the mask token. Dropout acts on the embedding, the attention weights and both residual branches.

\subsection{Training}

AdamW with $\beta = (0.9, 0.95)$, weight decay 0.1, gradient clipping at 1.0, linear warm-up over 2\% of the steps and a cosine schedule closed on the run's own length, batch 64. Each training budget is its own run, because intermediate checkpoints of a long cosine run are undertrained for the budget they represent \citep[Appendix B]{hoffmann2022chinchilla}. Training runs in fp16 with fp32 master weights and gradient scaling \citep{micikevicius2017mixed}, the loss in fp32 over the logits, and the attention kernel fixed to the same implementation on every machine. The budget counts applied steps only, and a step skipped by the scaler repeats the same batch, so the two modes see the same data in the same order.

\emph{Adapting to another domain.} Measure the throughput of the new machine before planning the runs, and keep the precision checks of \Cref{sec:checks} active. The learning rate and dropout must be searched again on the new corpus, as was done for corpus B.

\subsection{Generation}

Generation is ancestral. The autoregressive mode samples token by token. Masked diffusion starts from a fully masked window and, in $T$ steps, unmasks a random subset of positions, with the categorical draw in float64 \citep{zheng2024timeagnostic}. Unmasking by confidence, which replaces the draw by an argmax, enters only as a labelled variant.

\section{Evaluating the model}\label{sec:evaluation}

A model is good only in relation to something and for some purpose. The evaluation therefore has three layers. Likelihood says how well the model predicts unseen text of the domain. References give that number a meaning, by placing the model between a classical floor and a larger public model. Tasks and the quality of the samples say whether the model is useful.

\subsection{Likelihood in bits per byte}

Both modes are evaluated in bits per byte on the validation partition, as a ratio of sums over documents,
\begin{equation}
\bpb = \frac{\sum_d \text{nats}_d}{\ln 2 \cdot \sum_d \text{bytes}_d},
\end{equation}
with the bytes counted on the normalised text. The unit does not depend on the tokenizer, so models with different tokenizers, including the reference models below, are measured on the same scale. For the autoregressive mode the nats are the exact negative log-likelihood. For masked diffusion they are the NELBO estimated with the number-of-masks form in fp32, every window evaluated once, which is an upper bound on the true value. The windows are disjoint, each restarts its context, and the padding is predicted and counted, a design that closes the length leak between the modes (\Cref{sec:checks}). These bits per byte are therefore not comparable with evaluations that slide a window over running text, as is usual in the literature.

The bound makes the comparison between the modes asymmetric, and the reading is fixed before any run. With $\Delta = \bpb_{\mathrm{dif}} - \bpb_{\mathrm{AR}}$, an interval below zero means that diffusion is better in likelihood. An interval above zero means that the autoregressive mode beats the diffusion bound, with no conclusion about the true likelihood of diffusion. An interval containing zero is inconclusive.

\subsection{Reference points}\label{sec:references}

Two references bracket the model. The floor is an n-gram model with interpolated modified Kneser--Ney smoothing \citep{chen1996smoothing} trained on the same training partition and tokens, the classical method for small corpora, which a neural model must beat to show that it learned more than counts. The package implements it in a few dozen lines, with the same didactic purpose as the hand-written BPE, and two tests guard it. Every conditional distribution sums to one within $10^{-12}$, in seen and unseen contexts, and the probabilities equal those of a second, direct implementation of the formulas. The order is chosen on validation from 3 to 6. On corpus A it is 5, although orders 4 to 6 differ by 0.0003 bits per byte, so the choice is immaterial. On corpus B it is 6, the end of the grid, where the value was still falling. Training takes seconds on one CPU core. The floor has two values, and both are reported. Scored on running text, as a reader who uses it as a tool would, it reaches 1.833 bits per byte on corpus A, and 1.038 on validation, 1.046 on test and 1.206 on the external laws of 2025 and 2026 on corpus B. Scored on the same windows of 256 tokens as the Transformer, with the context restarted and the padding predicted, it reaches 1.841 on corpus A and 1.066 on corpus B, and 1.234 on the external laws. The comparison with the Transformer always uses the second value, which is the one to cite as the floor of this protocol. The yardstick is a larger public language model pretrained on general Portuguese text and used only for evaluation, never for training, measured on the same validation text in bits per byte. TeenyTinyLlama \citep{correa2024teenytiny}, already adapted to a specialist Portuguese domain by \citet{di2024slim}, is a natural candidate. A small model built for the domain that comes close to the yardstick, or passes it, shows that domain data can substitute for size. The public yardstick must not have seen the evaluation text. Tucano draws on a legal corpus with federal bills, and the Common Crawl snapshot of early 2023 holds 191 pages of laws from 2019 to 2022 on the official site, so on corpus B the yardstick is measured on the external acts only. Three yardsticks are used, TeenyTinyLlama 160M, Tucano 160M and Tucano 630M \citep{correa2024teenytiny,correa2024tucano}, at pinned revisions. Before any number, their 16-bit logits were compared with fp32 on 64 items, with a mean absolute difference of 0.003 nats per token and a maximum of 0.13, and the measurement itself ran in fp32. Their tokenizer drops the ordinal indicators º and ª, frequent in legal text, which amounts to 0.40\% of the bytes of the continuation task. Charging 16 bits for each dropped character would move a yardstick by at most 0.03 bits per byte, which is reported rather than corrected.

\subsection{Tasks in the legal domain}\label{sec:tasks}

Likelihood on validation says how well the model predicts text like its training text. Usefulness is asked of text the model cannot have seen, so the tasks are built on an external set of federal acts published in 2025 and 2026, after the training laws and after every public model used here. It holds 1,094 acts, 300 ordinary laws, 27 complementary laws and 767 decrees, of the substantive and amending classes, grouped as in corpus B so that a decree that regulates a law falls in the group of that law, 855 groups in all. Every task is built by a script from the text alone, with no human labelling, and each item is scored by every model through the same rule, the log-probability of a candidate text given its context.

\paragraph{T1, filling in the blank.} A number written in the legal pair ``N (in words)'' or the number of a cross-reference such as ``art.~N'' is deleted from an article, and the model chooses among four candidates with the same number of tokens. Number distractors come from the numbers of the training text with the same grammatical gender, and cross-reference distractors from the other articles cited by the same act. The metric is accuracy, with chance at 0.25.

\paragraph{T2, inserted or own provision.} An amending law inserts provisions into another law, quoted in its own text, and also has provisions of its own. Given the head of an article, the model decides which of the two it is. The score is the difference in log-probability of the text after an insertion formula (``passa a vigorar com a seguinte redação'') and after a plain article heading, and the threshold is fitted on training items for balanced accuracy, the metric of the task, since only 313 of the 3,095 external items are inserted. T2 was preregistered as confirmatory and is read as exploratory, for the reason given in \Cref{sec:taskresults}.

\paragraph{T3, tampered text.} An article is altered by one of four operations, a legal term swapped for its usual counterpart, numbers permuted within the article, a cross-reference changed to another article cited in the act, or the contents of two subdivisions of the same kind, such as two items or two paragraphs, swapped under their labels, one item per operation that applies. The model must prefer the original over the tampered version, with chance at 0.5.

\paragraph{T4, continuation.} The head of an article is cut at half of its words, and the model scores the real second half given the first, in bits per byte. A sample generated by the model and compared with the real text by surface overlap would be a weaker score, since overlap rewards near-verbatim reproduction and misses meaning kept in other words, a tension documented for translation by large language models in \citet{weigang2025paradox}.

\paragraph{Scoring and filters.} The autoregressive mode, the n-gram and the public models score a candidate by the exact log-probability of its tokens. Masked diffusion scores it by the conditional estimator of LLaDA \citep{nie2025llada}, with 128 masks per item drawn with the number of masks stratified and the same masks for all candidates of an item, an upper bound on the negative log-likelihood like the NELBO. An item is discarded if it shares a span of 13 tokens with any training document, so that recitation cannot answer it, and if it does not fit the window of 256 tokens with its context. \Cref{tab:taskcounts} gives the counts.

\begin{table}[!htbp]
\centering
\caption{Items of the four tasks on the external acts of 2025 and 2026, and the share of groups that must favour the model for a sign test to reach power 0.8 under Holm's correction in its worst case. The T2 share is kept as preregistered, although its sign test turned out not to measure the balanced accuracy (\Cref{sec:taskresults}). T1 holds 308 cross-references and 29 numbers.}
\label{tab:taskcounts}
\small
\begin{tabular}{lrrc}
\toprule
Task & Items & Groups & Share of groups for power 0.8 \\
\midrule
T1 Filling in the blank & 337 & 130 & 0.655 \\
T2 Inserted or own provision & 3,095 (313 inserted) & 643 & 0.570 \\
T3 Tampered text & 1,185 & 469 & 0.581 \\
T4 Continuation & 2,251 & 615 & 0.571 \\
\bottomrule
\end{tabular}
\end{table}

\paragraph{Statistics.} The unit is the group, not the item, since the provisions of one act and of the acts that amend it are not independent. The statistic is the mean of the per-group values, which gives each group the same weight, with a 95\% interval by a wild cluster bootstrap with Rademacher weights on the group influences, in T2 after linearising the recall of each class. The model beats the floor on a task when a two-sided sign test over the groups rejects at 0.05 in its favour, with Holm's correction over the comparisons of each track \citep{dror2018hitchhiker}, and the analysis is repeated without the largest group. T1 is descriptive. The public models are placed by their intervals, with no pass or fail. The design, the tasks, the item hashes and the decision rule were frozen before any measurement on the external set, and two dated amendments, each frozen by its hash before any score on the external set was computed, enlarged the external set from 300 laws to the 1,094 acts and generated one T3 item per applicable operation, both for power, and removed 13 T2 items that did not fit the window together with the insertion formula. The decrees therefore entered by a rule fixed before the measurement, not by a choice made after it, and the results are reported by genre as well as in total.

\subsection{From generator to typed decider}\label{sec:decider}

A typed decider receives a text and returns a probability distribution over fixed options in one pass, without generating text. Reading T2 and T3 as decisions, the network trained here can serve as the encoder of one, which lets a reader build a decider from scratch with the same pipeline. The states of the last layer, averaged over the tokens of the item, feed a linear head over the options. The autoregressive mode reads with causal attention and the diffusion mode with bidirectional attention and no mask. The whole network is fine-tuned on the training items for 3 epochs with AdamW at its default settings, batch 16, a constant learning rate chosen in $\{10^{-5}, 3 \times 10^{-5}, 10^{-4}\}$ by 5-fold cross-validation grouped by act, and cross-entropy weighted by the inverse class frequency. In T3 each text is judged alone as original or tampered, since a decider sees one input. A temperature is fitted on the out-of-fold logits, calibration is measured by the expected calibration error over 15 bins, and the selective answer abstains below the lowest confidence whose out-of-fold accuracy reaches 95\%. The floor is TF-IDF of words and word pairs with balanced logistic regression, its regularisation chosen in the same cross-validation. The public reference is the multilingual Laya decider\footnote{\texttt{convaiinnovations/laya}, revision \texttt{55cf4c4}, encoder mmBERT-base with 322M parameters.}, fine-tuned by the same protocol and also read without fine-tuning.

\subsection{Quality of the samples}

The quality of generated text is measured at four levels. Level N1, valid words, is automatic, measured against the lexicon of the training partition, since a modern dictionary would mark the period spelling of corpus A as error. Levels N2 (local syntax), N3 (complete grammatical sentence) and N4 (paragraph coherence) need a blind human reading and a semantic judge validated against it, both preregistered and described in \Cref{app:reading}, and their results are not part of this version. Before any reading, a sample is discarded if it copies from training a span longer than the 99th percentile of the longest span shared between human test excerpts and training \citep{mccoy2021raven,carlini2022memorization}, and the discard rate is reported as a measure of memorisation. This version reports the level N1, the copy filter and two diagnostics. The novelty of word n-grams, for $n$ from 1 to 10, is the share of the n-grams of a sample absent from training, compared with human test excerpts. The generative perplexity of samples under an external model, here Tucano 630M, is reported beside the entropy of each sample, since a lower effective temperature lowers the first while diversity falls \citep{zheng2024timeagnostic}.

\subsection{Calibration}

A good model also knows how sure it is. The unit is the prediction of one token, every position given its prefix in the autoregressive mode and every masked position of one draw of $l$ in diffusion, over 50,000 validation predictions taken from windows in a fixed random order, with the padding after the document's \EOS{} left out. Confidence is the largest softmax probability, and a prediction is correct when it equals the target. A temperature $T = 1/\beta$ is fitted by Newton's method on the mean negative log-likelihood, which is convex in $\beta$, the temperature scaling of \citet{guo2017calibration}. The expected calibration error averages the gap between confidence and accuracy over 15 equal bins, the reliability diagram plots one against the other, and the area under the ROC curve measures how well confidence separates correct from wrong predictions. The selective answer accepts the predictions above a threshold, the lowest one whose accepted predictions on a calibration half reach a target accuracy by the lower 5\% quantile of a Beta distribution, and is then applied to the other half, so that a promise made on one half is checked on the other. The deletion curve takes passages of 50 test words and the first token of the next word, removes the context words in order of importance, the drop in the probability of the decision when each one is removed, and records the remaining probability relative to the full context, against the same curve with words removed at random. The definitions follow the material of a typed-decider audit\footnote{``Confidence Is Not a Promise'', frozen answers, code and proofs at \doi{10.5281/zenodo.23676316}.}.

\section{Comparing configurations reliably}\label{sec:protocol}

Choices such as the learning rate or the generation mode are made by comparing runs, and small models vary from seed to seed. The protocol below makes a difference between configurations trustworthy, and it applies unchanged to any domain.

\paragraph{Checks before results.} Each design decision, on the formulation, precision, tokenizer, corpora and budget, lists checks in order, each with a criterion fixed before the measurement and an action if it fails (\Cref{app:gates}). The checks of the formulation run on CPU with toy sizes and the package's own code. With $L \le 6$ and a small vocabulary, enumerating all orders of unmasking confirms that the network's distribution over sequences sums to one and that the NELBO bounds the exact negative log-likelihood, and two deliberately injected defects must fail the same test.

\paragraph{Preregistration.} Every set of runs that decides something is preceded by a file with its design and decision rule, frozen by its SHA-256 hash, and the analysis script refuses to run if the file changed. A change after the data is allowed only as a dated amendment that states what had been seen.

\paragraph{Hyperparameter search.} The same grid in both modes, learning rate in $\{3 \times 10^{-4}, 10^{-3}, 3 \times 10^{-3}\}$ by dropout in $\{0, 0.1, 0.2\}$, 2 seeds per configuration, at 16 and 256 epochs, chosen at the 256-epoch point, with every configuration reported (\Cref{app:search}), as \citet{dodge2019show} recommend. If the two best configurations differ by less than the seed spread, the more regularised one wins.

\paragraph{Minimum effect.} A difference smaller than $\Delta_{\min}$ is treated as irrelevant. It is set to a quarter of what the autoregressive mode gains between 4 and 16 epochs, which on a logarithmic axis is what it gains with about 41\% more training. Measured in the search and frozen before the pilot, it is $\Delta_{\min} = 0.0907$ bits per byte on corpus A, with a standard error of 0.0025.

\paragraph{Number of seeds.} A pilot of 10 seeds per mode at the decisive budget estimates the spread $\hat\sigma$ of the paired difference by seed and does not enter the final analysis. With $\sigma$ taken at the upper limit of its 95\% interval, $1.83\,\hat\sigma$ with 9 degrees of freedom, a paired $t$ with power 0.8 and $\alpha = 0.05$ gives the number of seeds, with a ceiling of 30. On corpus A the pilot gave $\hat\sigma = 0.0084$ and $\sigma_{\sup} = 0.0154$, hence 3 seeds.

\paragraph{Inference.} A two-level bootstrap resamples works and seeds \citep{sellam2021multiberts}. Linear mixed models combined with bootstrap resampling are an alternative that decomposes the variance of an evaluation into fixed effects, such as the method, and random effects, such as the phrasing of a prompt \citep{di2025immba}. Here the random factors are the seed and the work. Its coverage is checked in the pilot with 1,000 null replicates built by flipping the sign of the paired differences, and it must reach 0.93 for a nominal 95\% interval. If it does not, a sign-flip randomization test over seeds decides, and the claim is restricted to the evaluation set. On corpus A the coverage was 0.918, and the randomization test became the final analysis.

\paragraph{Amendments.} Two amendments changed the protocol on corpus A, and both are dated in the repository. On 26 October 2026, after the first precision check, a criterion that compared the gradient error of each tensor in diffusion with twice its error in the autoregressive mode was dropped, because the ratio of two errors near the rounding of fp16 has no absolute floor. The interaction between precision and mode was judged instead by the fp16 versus fp32 difference in bits per byte. This amendment came after the data and is recorded as such. On 27 October 2026, before any final run, the number of seeds rose from 3 to 6, because with 3 seeds the randomization test has 8 sign patterns and its smallest two-sided $p$ is 0.25, so it could not reject at 0.05 whatever the effect. The rule takes the larger of the power formula and the smallest $n$ with $2 \cdot 2^{-n} < 0.05$, and corpus B applies it from the start. The sensitivity analysis of \Cref{sec:sensitivity} was preregistered before its runs, and one dated amendment corrected a count cited in it. The preregistration stated that no step had been skipped by the gradient scaler, from a script that read the count under a wrong field name, and the count from the run summaries is 123 in the 104 runs of the comparison, at most 10 in one run.

\paragraph{Scope of the claims.} The training corpus has far fewer tokens than the compute-optimal target for the network \citep{hoffmann2022chinchilla}, so training repeats the same text many times, and this regime is part of the design. The budget grid has five points, 1, 4, 16, 64 and 256 epochs, with 5 seeds at the four descriptive points and the hypothesis test at 256 epochs only. Every claim holds for unseen chapters of the works of the corpus up to 256 epochs.

\section{Worked example}\label{sec:results}

\subsection{Throughput and precision}

On the reference machine, with $V = 4096$, $L = 256$ and batch 64, training in fp16 reached a model FLOPs utilisation of about 0.41 in the autoregressive mode and 0.38 in diffusion, an approximate figure, since the peak of 209.5 TFLOPS behind it comes from secondary sources. In fp32 the rates were 448k and 430k tokens per second. Each run of 256 epochs took 11,120 applied steps, with 2 or 3 steps skipped by the gradient scaler. To confirm that half precision does not change the result, 3 seeds in fp16 were compared with 3 in fp32 per mode at 16 epochs. The 95\% interval of the difference in bits per byte, paired by seed, was $[-0.0143, 0.0153]$ in the autoregressive mode and $[-0.0094, 0.0048]$ in diffusion, both inside $\pm \Delta_{\min}/4 = \pm 0.0227$, so training proceeded in fp16.

\subsection{Likelihood across training budgets}

\begin{figure}[!htbp]
\centering
\includegraphics[width=0.7\linewidth]{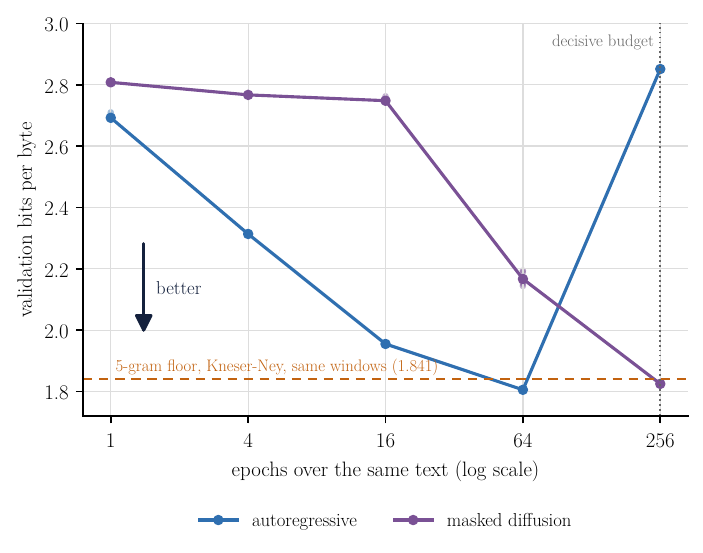}
\caption{Validation bits per byte against epochs over the same text, corpus A. Points are seeds (5 at the descriptive budgets, 6 at the decisive budget, dotted line), lines join the means. The dashed line is the 5-gram floor scored on the same windows of 256 tokens. The diffusion values are upper bounds.}
\label{fig:epochs}
\end{figure}

\Cref{fig:epochs} and \Cref{tab:grid} show the validation bits per byte at each budget, with the hyperparameters chosen in the search. The autoregressive mode is better at 1, 4, 16 and 64 epochs, by 0.115 to 0.793 bits per byte. At 256 epochs the order reverses. Masked diffusion reaches 1.825 against 2.851, $\Delta = -1.026$, and all six seed differences lie between $-1.029$ and $-1.023$. The sign-flip randomization test gives $p = 0.031$, the smallest value attainable with 6 seeds, so by the preregistered reading diffusion is better in likelihood at this budget, a claim restricted to the evaluation set. The magnitude of $\Delta$ is about eleven times $\Delta_{\min}$. The bootstrap interval, $[-1.099, -0.954]$, is given for scale only, since its coverage fell below the preregistered minimum.

\begin{table}[!htbp]
\centering
\caption{Mean validation bits per byte per mode and paired difference $\Delta = \bpb_{\mathrm{dif}} - \bpb_{\mathrm{AR}}$, corpus A. Intervals from the two-level bootstrap, whose coverage in the pilot was 0.918. At 256 epochs the decision rests on the randomization test.}
\label{tab:grid}
\small
\begin{tabular}{rcccccc}
\toprule
Epochs & Seeds & Autoregressive & Diffusion & $\Delta$ & 95\% interval & Randomization $p$ \\
\midrule
1 & 5 & 2.692 & 2.808 & 0.115 & $[0.088, 0.161]$ & \\
4 & 5 & 2.314 & 2.767 & 0.453 & $[0.438, 0.473]$ & \\
16 & 5 & 1.955 & 2.748 & 0.793 & $[0.775, 0.811]$ & \\
64 & 5 & 1.806 & 2.167 & 0.361 & $[0.332, 0.395]$ & \\
256 & 6 & 2.851 & 1.825 & $-1.026$ & $[-1.099, -0.954]$ & 0.031 \\
\bottomrule
\end{tabular}
\end{table}

\subsection{What repetition does to each mode}\label{sec:repetition}

The observations in this subsection are descriptive, since only the 256-epoch budget carries a hypothesis test, and they hold for this model and these corpora. The autoregressive mode learns faster. It improves at every budget up to 64 epochs and then degrades sharply, ending at 256 epochs worse than at 4. Masked diffusion barely moves from 1 to 16 epochs, 2.808 to 2.748, and then improves at every budget up to the end of the grid, so the grid does not show where it stops improving. The comparison a reader is most likely to need is between the best point of each mode, 64 epochs for the autoregressive mode and 256 for diffusion, and it was not preregistered. A Welch TOST with margin $\Delta_{\min}$, added after the decisive analysis and declared as such, compares the 5 autoregressive seeds with the 6 of diffusion. On validation, where the best points were chosen, diffusion minus autoregressive is 0.019 bits per byte on corpus A, 90\% interval $[0.016, 0.023]$, and 0.069 on corpus B, $[0.064, 0.074]$. On test, the less biased reading because no choice was made there, it is 0.060 on corpus A, $[0.056, 0.064]$, which is 66\% of $\Delta_{\min}$, and 0.061 on corpus B, $[0.058, 0.065]$, 47\% of $\Delta_{\min}$. Both tests reject at 0.05. Read asymmetrically, as fixed in \Cref{sec:evaluation}, this says that the diffusion bound is not above the exact autoregressive likelihood by more than $\Delta_{\min}$, and that the autoregressive mode is ahead in point estimate, with every interval above zero. It says nothing about equivalence in true likelihood, since the gap between the bound and the true value of diffusion was not measured. For a reader with a small corpus, the practical consequence is that the training budget must be chosen per mode, since the best budget of one is a poor budget for the other.

\begin{figure}[!htbp]
\centering
\includegraphics[width=\linewidth]{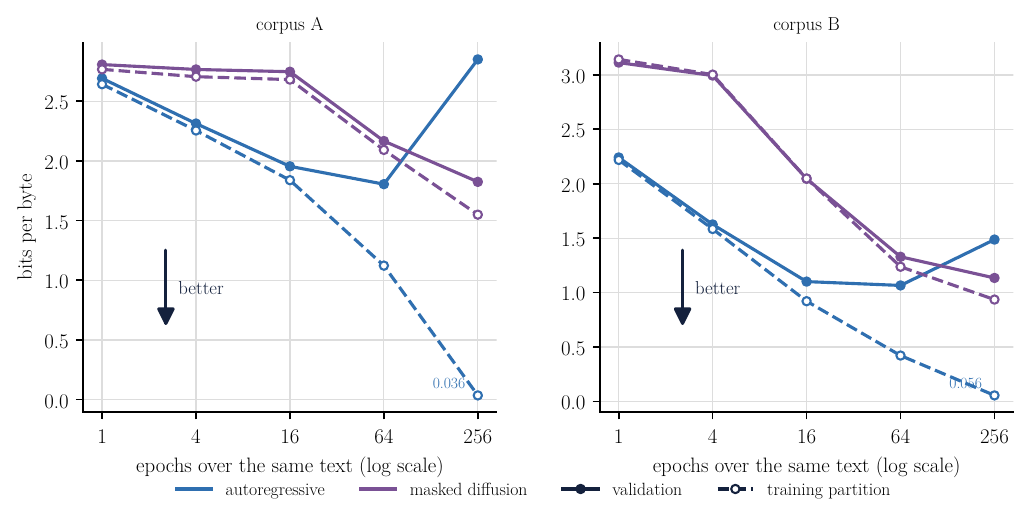}
\caption{Bits per byte on the training partition (dashed, open points) and on validation (solid) across the budget grid, mean over seeds. At 256 epochs the autoregressive mode nearly reproduces its training text, and the gap to validation is the overfitting of \Cref{fig:epochs}. The diffusion values are upper bounds.}
\label{fig:memorisation}
\end{figure}

Scoring the training partition shows what the degradation is (\Cref{fig:memorisation}). At 256 epochs the autoregressive mode reaches 0.036 bits per byte on its own training text against 2.851 on validation, so it has memorised the training chapters almost exactly, while diffusion reaches 1.550 on training against 1.825 on validation. Corpus B repeats the pattern, 0.056 against 1.487 for the autoregressive mode and 0.936 against 1.135 for diffusion. The samples point the same way, with about 30 samples per budget. At 256 epochs the copy filter discards 1 of 29 autoregressive samples for a span of more than 7 words copied from training, and it discards none of the diffusion samples at any budget.

The pattern agrees in direction with what \citet{prabhudesai2025data} report at larger scale, with 100M unique tokens, where diffusion overtakes the autoregressive mode when data are scarce and repeated, and the half-life of data reuse is about 500 epochs for diffusion against about 31 for the autoregressive mode. \citet{ni2025superdata} report results in the same direction. On the autoregressive side, \citet{muennighoff2023scaling} find that up to 4 epochs of repeated data change the loss little compared with unique data and that the value of further repetition decays to zero, and \citet{xue2023repeat} find that repeated epochs degrade the model and that dropout was the one regulariser among those they tried that proved effective against it. \citet{nie2024scaling}, in the single-epoch regime, estimate that masked diffusion needs about 16 times the compute of the autoregressive mode to match it, which agrees in direction with the 1-epoch point here. \citet{prabhudesai2025data} attribute the tolerance of diffusion to repetition to masking acting as data augmentation, since each pass over a sequence masks different positions. That mechanism was not tested here.

A control at equal compute points the same way. The trivial byte tokenizer of the contract, with 260 ids, was trained in the autoregressive mode with the hyperparameters chosen for that mode, over 3 seeds, and received the total FLOPs of the BPE runs at the 16-epoch and the 256-epoch budgets. A BPE token covers 3.56 bytes of training text on average, so at equal compute the byte model passes over the text fewer times, 0.28 to 0.32 of the BPE epochs. A window of 256 bytes holds less text than a window of 256 BPE tokens, about 911 bytes, so a second arm uses a window of 911 bytes to equalise the context. At the 16-epoch budget the byte model is worse by 0.80 to 1.41 bits per byte. At the 256-epoch budget it is better by about 1.0 bit per byte with either window, reaching 1.84 to 1.85, close to the best autoregressive value of the grid (\Cref{tab:bytes}). A reading consistent with the grid is that matching compute made the byte model repeat the text 72 to 81 times instead of 256, and so overfit less. This was not measured separately, and a property of byte-level modelling as such cannot be excluded.

\begin{table}[!htbp]
\centering
\caption{Byte-level control at equal compute, autoregressive mode, corpus A, 3 seeds. Epochs are passes over the training text, $L$ is the window in symbols, and the difference to BPE is paired by seed, with a 95\% interval by $t$ on 2 degrees of freedom.}
\label{tab:bytes}
\small
\begin{tabular}{lcccc}
\toprule
Tokenizer & Epochs & $L$ & Bits per byte & Difference to BPE \\
\midrule
BPE, $V = 4096$ & 16 & 256 & 1.954 & \\
Bytes, $V = 260$ & 5.1 & 256 & 2.751 & $+0.797$ $[+0.760, +0.834]$ \\
Bytes, $V = 260$ & 4.5 & 911 & 3.368 & $+1.414$ $[+1.405, +1.422]$ \\
\midrule
BPE, $V = 4096$ & 256 & 256 & 2.849 & \\
Bytes, $V = 260$ & 81.3 & 256 & 1.849 & $-1.000$ $[-1.024, -0.976]$ \\
Bytes, $V = 260$ & 71.8 & 911 & 1.840 & $-1.010$ $[-1.026, -0.993]$ \\
 
\bottomrule
\end{tabular}
\end{table}

Two features of the protocol shape this picture. The hyperparameters of each mode were chosen at the 256-epoch budget and held fixed across the grid, so the descriptive budgets do not show each mode at its best for that budget. At 16 epochs, the best autoregressive configuration of the search reached 1.851, against 1.955 with the chosen one. And the configuration chosen for the autoregressive mode sits at the corner of the grid, the lowest learning rate and the highest dropout (\Cref{app:search}), so a stronger regulariser outside the grid might have reduced its overfitting at 256 epochs, and dropout is the one \citet{xue2023repeat} found effective. The result at 256 epochs is conditional on the grid that was preregistered.

\subsection{The specialist domain}

Corpus B repeats the protocol from the start, with a hyperparameter search of its own, preregistered under the hash \texttt{22e3c824}. The search again chose the corner of the grid for the autoregressive mode, learning rate $3 \times 10^{-4}$ and dropout 0.2, and learning rate $10^{-3}$ with dropout 0.1 for diffusion. The minimum effect is $\Delta_{\min} = 0.1304$ bits per byte, with a standard error of 0.0005. The pilot asked for 3 seeds, the rule of the smallest attainable $p$ raised them to 6, and the bootstrap failed its coverage check again, at 0.919, so the randomization test decides.

\begin{figure}[!htbp]
\centering
\includegraphics[width=0.7\linewidth]{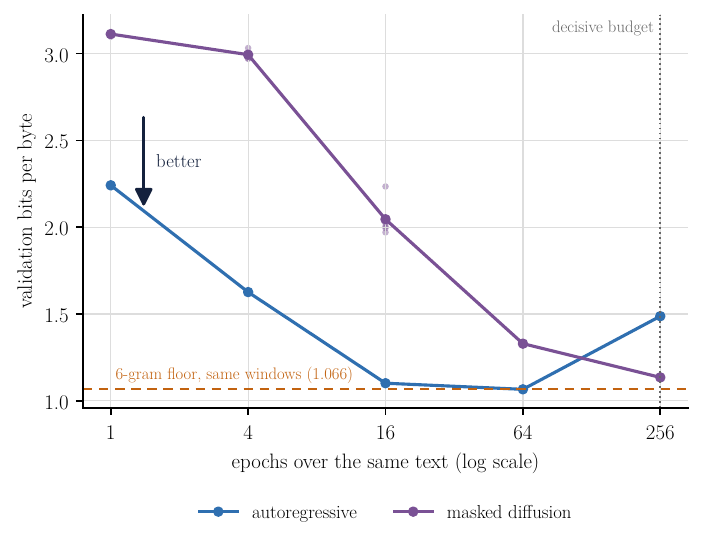}
\caption{Validation bits per byte against epochs over the same text, corpus B, $V = 1024$. Points are seeds, lines join the means, and the dashed line is the 6-gram floor on the same windows of 256 tokens. The diffusion values are upper bounds.}
\label{fig:epochsB}
\end{figure}

The picture of corpus A repeats (\Cref{fig:epochsB}, \Cref{tab:gridB}). The autoregressive mode is better up to 64 epochs, where it reaches its best point, 1.066 bits per byte, and degrades to 1.487 at 256. Masked diffusion improves at every budget and reaches 1.135 at 256 epochs, where $\Delta = -0.353$, with all six seeds on the same side and the randomization test at its floor, $p = 0.031$. By the preregistered reading diffusion is better in likelihood at this budget, by 2.7 times $\Delta_{\min}$. At their best points the order is the other way, the autoregressive mode ahead by 0.069, below $\Delta_{\min}$.

\begin{table}[!htbp]
\centering
\caption{Mean validation bits per byte per mode and paired difference $\Delta = \bpb_{\mathrm{dif}} - \bpb_{\mathrm{AR}}$, corpus B. Intervals from the two-level bootstrap, whose coverage in the pilot was 0.919. At 256 epochs the decision rests on the randomization test.}
\label{tab:gridB}
\small
\begin{tabular}{rcccccc}
\toprule
Epochs & Seeds & Autoregressive & Diffusion & $\Delta$ & 95\% interval & Randomization $p$ \\
\midrule
1 & 5 & 2.242 & 3.113 & 0.871 & $[0.839, 0.911]$ & \\
4 & 5 & 1.626 & 2.994 & 1.368 & $[1.323, 1.421]$ & \\
16 & 5 & 1.101 & 2.046 & 0.945 & $[0.864, 1.053]$ & \\
64 & 5 & 1.066 & 1.329 & 0.263 & $[0.189, 0.351]$ & \\
256 & 6 & 1.487 & 1.135 & $-0.353$ & $[-0.509, -0.188]$ & 0.031 \\
\bottomrule
\end{tabular}
\end{table}

The change of domain is itself checked. A script compares the configurations of corpus A and corpus B and confirms that only the configuration and the domain adapter differ. The whole notebook ran on corpus B, selected by a single environment variable, in 81.8\,s on CPU with the short settings of the test.

\subsection{Confirmation on the test partition}\label{sec:test}

\begin{figure}[!htbp]
\centering
\includegraphics[width=\linewidth]{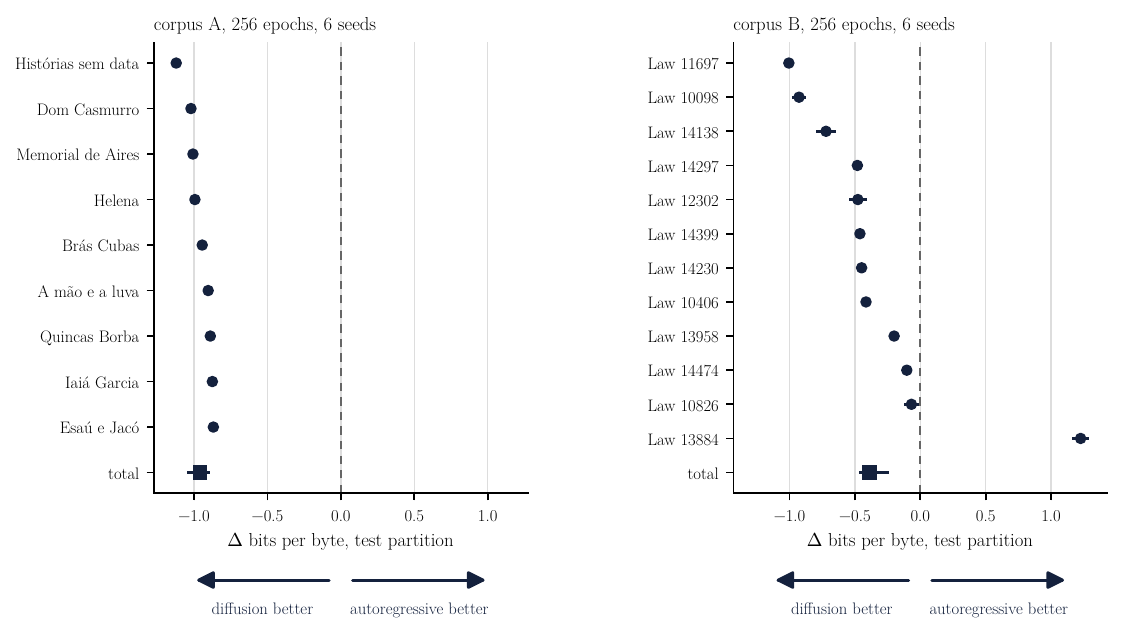}
\caption{$\Delta = \bpb_{\mathrm{dif}} - \bpb_{\mathrm{AR}}$ on the test partition at 256 epochs, per work of corpus A and per group of corpus B, mean of the 6 seeds with a $t$ interval, and the total with the two-level bootstrap interval. Left of zero, diffusion is better in likelihood.}
\label{fig:test}
\end{figure}

The comparison was decided on validation, and the test partition, on which the models were scored only after the decisive analysis, checks it once. On corpus A at 256 epochs, $\Delta = -0.961$ on test, with the six seed differences between $-0.968$ and $-0.947$, the randomization test at its floor, $p = 0.031$, and a bootstrap interval of $[-1.034, -0.901]$ given for scale. All nine works fall on the side of diffusion, from $-0.868$ for \emph{Esaú e Jacó} to $-1.122$ for \emph{Histórias sem data} (\Cref{fig:test}), and a sign test over the works, descriptive, gives a two-sided $p$ of 0.0039. Proper names, which cross the blocks of a work, cover 2.83\% of the test bytes, and leaving their tokens out of both the nats and the bytes moves $\Delta$ to $-0.977$, so the result does not rest on names seen in training. On corpus B, $\Delta = -0.389$ on test, $[-0.457, -0.253]$, $p = 0.031$, with 11 of the 12 groups below zero, a sign test $p$ of 0.0063. The exception is the group of Law 13,884, 0.6\% of the test bytes, where the autoregressive mode is ahead by 1.226, and the group of the Civil Code, 53\% of the test bytes, sits at $-0.415$.

\subsection{Sensitivity of the comparison}\label{sec:sensitivity}

\begin{figure}[!htbp]
\centering
\includegraphics[width=\linewidth]{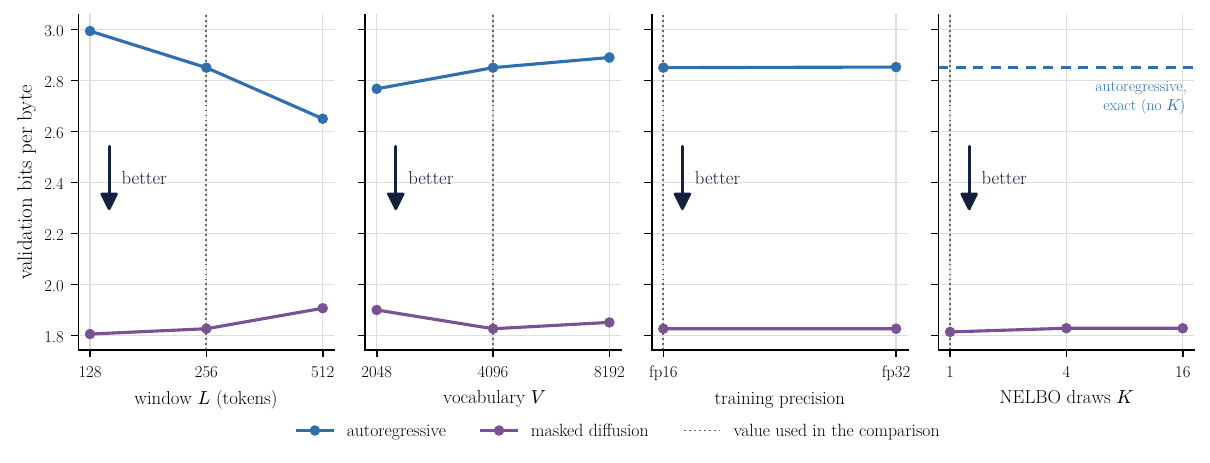}
\caption{Validation bits per byte at 256 epochs on corpus A with one factor changed at a time, seeds 300 to 302 as points and their mean as a line. The dotted line marks the value used in the comparison. $K$ applies to the diffusion bound only, and the dashed line is the exact autoregressive value.}
\label{fig:sensitivity}
\end{figure}

The comparison at 256 epochs fixes choices that a reader might make differently. A preregistered sensitivity analysis on corpus A changed one at a time, with seeds 300 to 302 paired with the decisive runs, and is descriptive (\Cref{fig:sensitivity}). With windows of 128 and 512 tokens, $\Delta$ was $-1.189$ and $-0.744$, against $-1.024$ for these seeds at 256. With vocabularies of 2,048 and 8,192, it was $-0.868$ and $-1.040$, the larger one although 12\% of its tokens never occur in the held-out text. Training in fp32 gave $-1.027$, with the autoregressive mode at 2.853 against 2.850 in fp16. The sign never changed, and the smallest magnitude was 8.2 times $\Delta_{\min}$. The autoregressive mode moved more than diffusion, from 2.650 to 2.994 against 1.805 to 1.906, which agrees with an overfitting that depends on how much text each step sees, a reading not tested separately.

The number $K$ of draws per window changes the estimate of the diffusion bound, not its expectation. Over three evaluation seeds per checkpoint, the standard deviation of a single evaluation was 0.034 bits per byte with $K = 1$, 0.0024 with $K = 4$ and 0.0014 with $K = 16$, and $\Delta$ was $-1.037$, $-1.022$ and $-1.023$. The budget check required a Monte Carlo error below $\Delta_{\min}/4 = 0.0227$, and that criterion was not computed before the pilot. Measured now, a single evaluation with $K = 1$ is above it and the mean of the six decisive seeds, about 0.014, below it, with no effect on the reading. A reader who evaluates single runs should use $K \ge 4$.

The number of generation steps trades time for quality. With 16, 64 and 256 steps, the lexicon rate of the diffusion samples rose from 0.865 to 0.876 and 0.881, the share of their word trigrams attested in training from 0.092 to 0.103 and 0.112, and the time per batch of 32 samples from 7 to 28 and 112 seconds. The autoregressive samples of the same seeds reach 0.952 and 0.234 in about one second, and human excerpts 0.962 and 0.178. Counting the budget by applied or by nominal steps differs by at most 3 of 11,120 steps on corpus A and 10 of 25,632 on corpus B, all at the end of the cosine schedule, and was not run separately.

\subsection{Why large models are autoregressive}\label{sec:whyar}

If diffusion wins at 256 epochs on both corpora, a reader may fairly ask why the large language models in wide use are autoregressive. Part of the answer is in the grid. At its own best budget the autoregressive mode is not worse here (\Cref{sec:repetition}), so what diffusion offers in this regime is tolerance to repetition, not a better best point. Large models are trained in the opposite regime, reading far more unique text than they could repeat, with compute as the binding limit, and there \citet{nie2024scaling} estimate that masked diffusion needs about 16 times the compute of the autoregressive mode to match it, which the 1-epoch points of both grids echo, with the autoregressive mode ahead by 0.115 and 0.871 bits per byte. Causal attention also lets the autoregressive mode cache the keys and values of past tokens, which plain masked diffusion cannot do. Block variants recover the cache \citep{arriola2025bd3lm}, and the speed-up reported for LLaDA 8B over LLaMA 3 8B with its cache is modest, 1.5 to 1.8 times at batch size 1 \citep[Appendix B.7]{nie2025llada}. And the autoregressive likelihood is exact where diffusion gives a bound. Why a given company keeps its choice is not observable from this work.

The reader of this tutorial is in the regime of the grid, little text read many times, where the choice of budget matters more than the choice of mode. Finding the best point of a mode means training the budget grid, one run per budget with the cosine schedule closed on its own length, and even then the grid locates that point only to within a factor of 4 in epochs. A reader who can pay for the grid can stop the autoregressive mode at its best point. A reader who cannot is better served by diffusion, which did not degrade anywhere in the grid, while the autoregressive mode lost ground on both corpora between 64 and 256 epochs.

\subsection{Is the model good?}

On corpus A the honest answer, by likelihood alone, is barely. The 5-gram floor, trained in a few seconds of CPU and scored on the same windows of 256 tokens, reaches 1.841 bits per byte on validation (\Cref{fig:epochs}). The autoregressive Transformer is worse than the floor at 1, 4 and 16 epochs, crosses it between 16 and 64, and at its best point, 1.806 at 64 epochs, beats it by 0.035, less than half of $\Delta_{\min}$. The diffusion mode at its best point, 1.825 at 256 epochs, beats it by 0.016. A reader who trains a small model on a few hundred thousand tokens and compares it only with other neural runs would not see this. The comparison with the floor is the first check of usefulness, and it costs seconds.

\begin{figure}[!htbp]
\centering
\includegraphics[width=\linewidth]{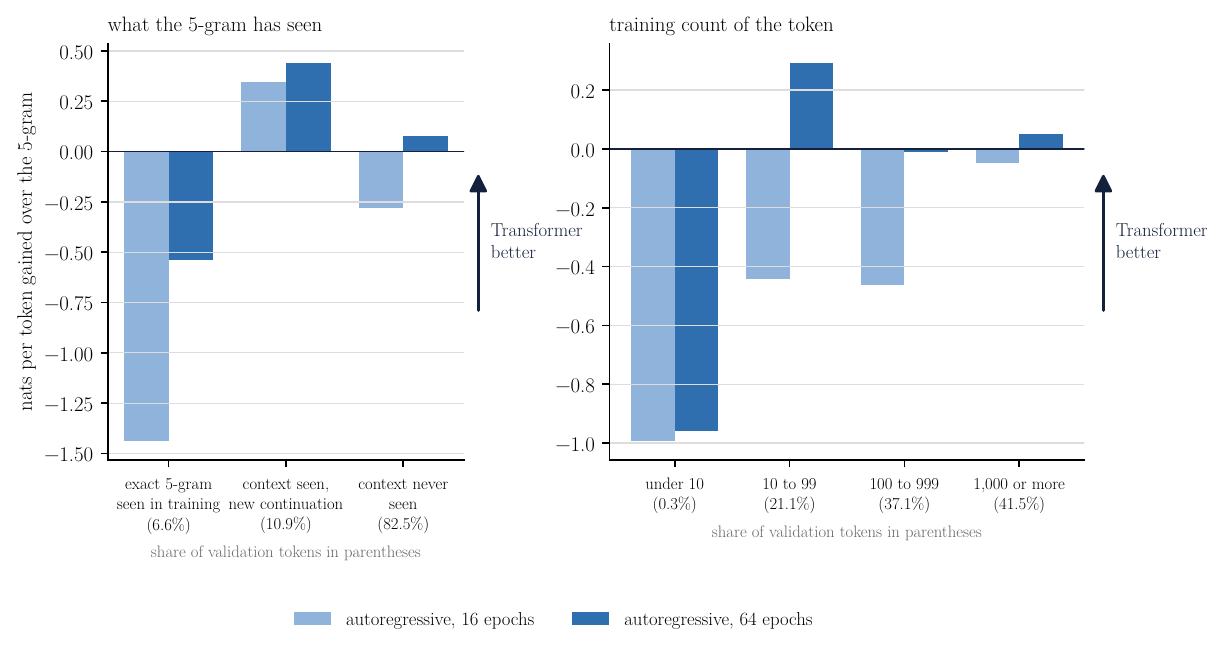}
\caption{Where the autoregressive Transformer beats the 5-gram floor on the validation of corpus A, in nats per token, at 16 and 64 epochs. Left, by what the 5-gram knows of the context. Right, by how often the predicted token occurs in training. Above zero the Transformer is better.}
\label{fig:where}
\end{figure}

\begin{figure}[!htbp]
\centering
\includegraphics[width=0.7\linewidth]{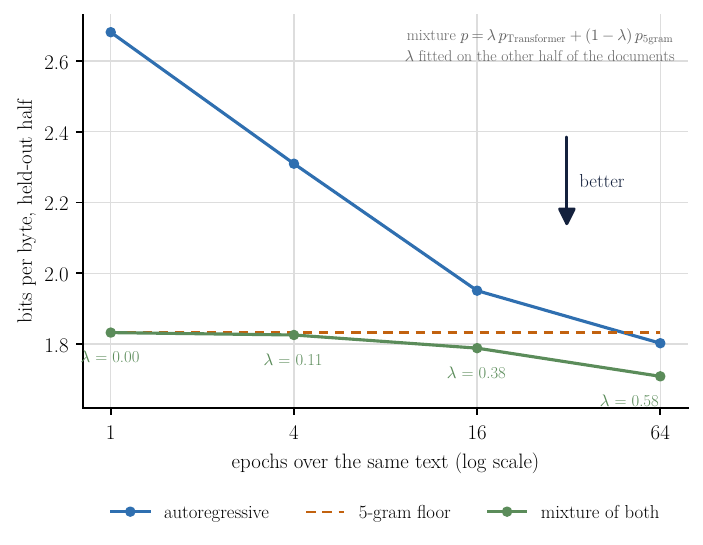}
\caption{Bits per byte on the odd validation documents of corpus A for the autoregressive Transformer, the 5-gram floor and their mixture, with the mixing weight $\lambda$ fitted on the even documents. The weight of the Transformer grows with training.}
\label{fig:mixture}
\end{figure}

The floor is a single number, though, and the two models are not good at the same things. Scoring every validation token with both, on the same windows, shows where the Transformer earns its place (\Cref{fig:where}). Where the exact sequence of five tokens already occurs in training, 6.6\% of the tokens, the n-gram wins by 0.54 nats per token, since it recalls repetitions verbatim. Where the context was seen with another continuation, 10.9\% of the tokens, the Transformer wins by 0.44 nats per token, which is generalisation from a known context to a new continuation. In the remaining 82.5\%, contexts the n-gram never saw, the Transformer wins by 0.08. By frequency, the gain concentrates on tokens seen 10 to 99 times in training. The two models fail in different places, and mixing their probabilities, with the weight fitted on half of the validation documents and measured on the other half, reaches 1.709 bits per byte at 64 epochs, against 1.802 for the Transformer and 1.832 for the n-gram on the same half (\Cref{fig:mixture}). The gain of the mixture over the Transformer, 0.093, is the size of $\Delta_{\min}$. For a home-built model, the practical lesson is to keep the n-gram and mix, as language modelling did before neural networks and as \citet{liu2024infinigram} do at the scale of trillions of tokens.

\subsection{What the samples show}\label{sec:samples}

\begin{figure}[!htbp]
\centering
\includegraphics[width=0.7\linewidth]{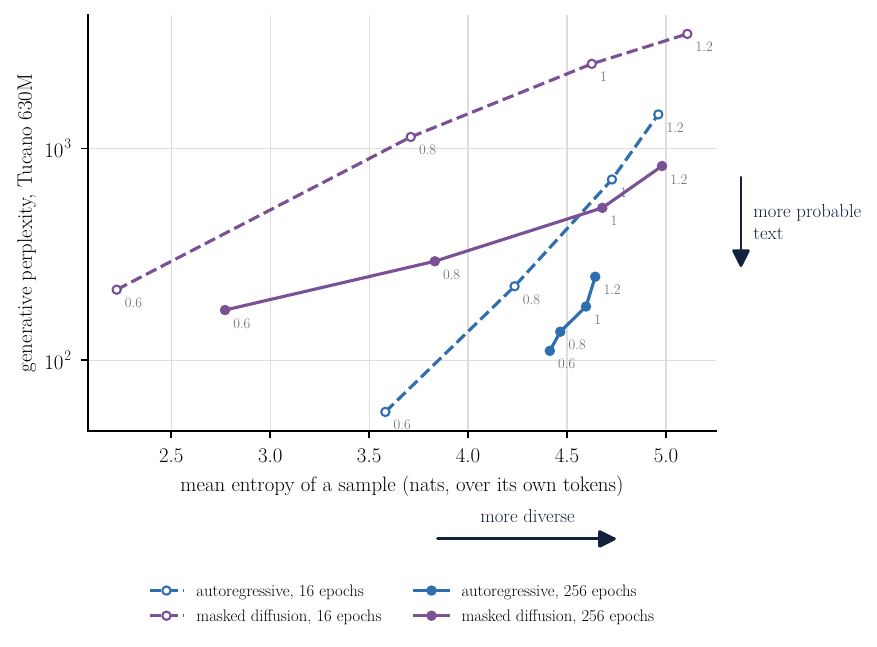}
\caption{Mean entropy of a sample over its own tokens against its generative perplexity under Tucano 630M, at sampling temperatures 0.6 to 1.2 (labels), 64 samples per seed and 3 seeds, corpus A. Dashed, 16 epochs. Solid, 256 epochs.}
\label{fig:entropy}
\end{figure}

On level N1 the autoregressive samples reach the lexicon rate of human text, 0.941 at 64 epochs and 0.950 at 256 against 0.962 for human test excerpts of the same length, within the preregistered 5 points from 64 epochs on. Diffusion reaches 0.808 and 0.889 and does not meet N1 anywhere in the grid. The novelty of word n-grams at 256 epochs shows two opposite departures from human text. The autoregressive samples reuse more short sequences from training, with 31\% of their word bigrams absent from training against 40\% for human excerpts, and diffusion invents more, 48\%. Generative perplexity under Tucano 630M moves with temperature in both modes as expected (\Cref{fig:entropy}). At temperature 1 and 256 epochs it is 179 for the autoregressive mode and 525 for diffusion, at nearly the same entropy, 4.60 and 4.68 nats per token. The lower value of the autoregressive mode agrees with its memorisation, and Machado's prose is in the public domain and may be in the pretraining data of the scorer, so a low generative perplexity here does not separate quality from recall. No decision rests on these diagnostics, and the levels N2 to N4 await the blind reading.

\subsection{Calibration and deletion}\label{sec:calresults}

\begin{figure}[!htbp]
\centering
\includegraphics[width=\linewidth]{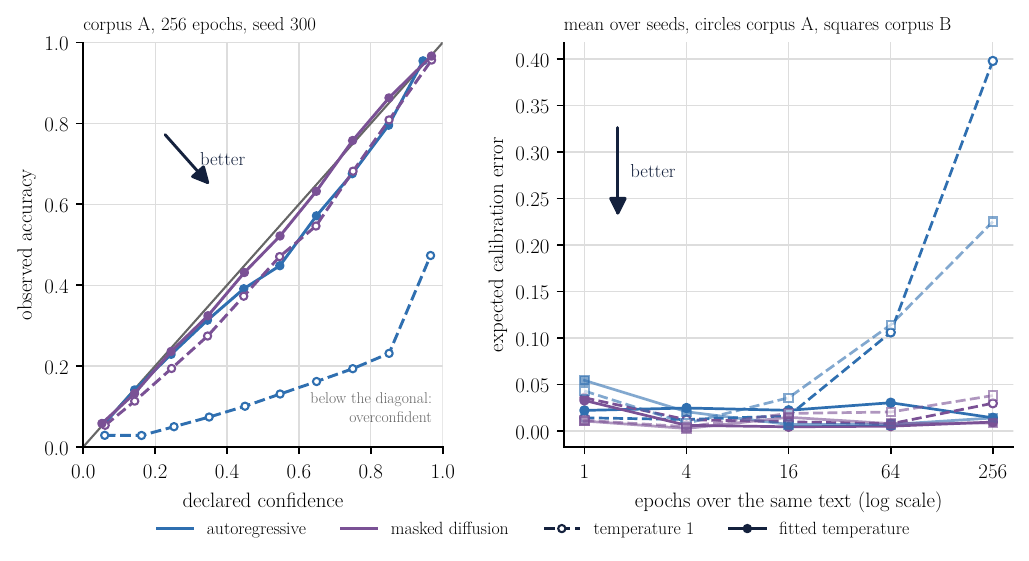}
\caption{Left, reliability of the first decisive seed of corpus A at 256 epochs, at temperature 1 (dashed, open points) and at the fitted temperature (solid). Right, expected calibration error across the budget grid, mean over seeds, corpus A (circles) and corpus B (squares, lighter).}
\label{fig:calibration}
\end{figure}

Calibration improves with training and depends on the mode (\Cref{tab:calibration}, \Cref{fig:calibration}). The autoregressive mode is well calibrated at temperature 1 up to 16 epochs, with an expected calibration error of 0.012 to 0.016 on corpus A, and becomes overconfident as it memorises, 0.106 at 64 epochs and 0.398 at 256, where the fitted temperature, 2.19, brings it back to 0.014. Diffusion stays below 0.04 at every budget of both corpora at temperature 1. In the first budgets, the temperature that minimises the negative log-likelihood can raise the calibration error, from 0.014 to 0.022 at one epoch, since the two criteria differ. Confidence also becomes a better detector of correct predictions with training, the area under the ROC curve rising from 0.65 to 0.79 in the autoregressive mode and from 0.47 to 0.83 in diffusion on corpus A. The selective answer at a 90\% target shows what a promise is worth. On corpus A it covers 1.3\% of the predictions in the autoregressive mode and 5\% in diffusion at 256 epochs, and the promise made on one half holds on the other in 4 of 6 and 6 of 6 seeds. On corpus B, a more repetitive text, it covers up to 49\% in the autoregressive mode. Diffusion finds no threshold that reaches the target before 64 epochs on corpus A and before 16 on corpus B.

\begin{table}[!htbp]
\centering
\caption{Calibration and deletion per budget, mean over the seeds of each budget (5, or 6 at 256 epochs), on validation predictions for calibration and test passages for deletion. ECE at temperature 1 and at the fitted temperature $T$, ROC AUC of confidence as a detector of correct predictions, selective answer at a 90\% target (coverage and seeds whose promise held on the held-out half), and area under the deletion curve by importance and at random.}
\label{tab:calibration}
\scriptsize
\begin{tabular}{lllcccccccc}
\toprule
 & & & & \multicolumn{2}{c}{ECE} & & \multicolumn{2}{c}{Selective at 0.9} & \multicolumn{2}{c}{Deletion AUC} \\
\cmidrule(lr){5-6}\cmidrule(lr){8-9}\cmidrule(lr){10-11}
Corpus & Mode & Epochs & $T$ & $T = 1$ & fitted & ROC AUC & Coverage & Kept & Importance & Random \\
\midrule
A & AR & 1 & 0.95 & 0.014 & 0.022 & 0.647 & 0.000 & 0/5 & 0.562 & 0.811 \\
A & AR & 4 & 0.96 & 0.012 & 0.025 & 0.748 & 0.000 & 0/5 & 0.234 & 0.702 \\
A & AR & 16 & 0.98 & 0.016 & 0.022 & 0.747 & 0.016 & 5/5 & 0.163 & 0.669 \\
A & AR & 64 & 1.18 & 0.106 & 0.030 & 0.779 & 0.029 & 5/5 & 0.102 & 0.501 \\
A & AR & 256 & 2.19 & 0.398 & 0.014 & 0.792 & 0.013 & 4/6 & 0.080 & 0.328 \\
\addlinespace
A & Diff. & 1 & 1.07 & 0.036 & 0.033 & 0.471 & 0.000 & 0/5 & 0.997 & 0.999 \\
A & Diff. & 4 & 1.03 & 0.012 & 0.006 & 0.501 & 0.000 & 0/5 & 1.000 & 1.000 \\
A & Diff. & 16 & 1.02 & 0.010 & 0.005 & 0.528 & 0.000 & 0/5 & 0.929 & 0.963 \\
A & Diff. & 64 & 1.01 & 0.008 & 0.005 & 0.793 & 0.014 & 5/5 & 0.248 & 0.704 \\
A & Diff. & 256 & 1.08 & 0.030 & 0.010 & 0.831 & 0.050 & 6/6 & 0.149 & 0.605 \\
\midrule
B & AR & 1 & 0.82 & 0.043 & 0.054 & 0.755 & 0.022 & 5/5 & 0.289 & 0.704 \\
B & AR & 4 & 0.95 & 0.008 & 0.021 & 0.842 & 0.139 & 3/5 & 0.192 & 0.663 \\
B & AR & 16 & 1.11 & 0.036 & 0.007 & 0.890 & 0.398 & 4/5 & 0.128 & 0.581 \\
B & AR & 64 & 1.48 & 0.114 & 0.007 & 0.902 & 0.493 & 3/5 & 0.094 & 0.501 \\
B & AR & 256 & 2.23 & 0.225 & 0.014 & 0.903 & 0.492 & 6/6 & 0.085 & 0.438 \\
\addlinespace
B & Diff. & 1 & 1.05 & 0.012 & 0.011 & 0.500 & 0.000 & 0/5 & 0.965 & 0.977 \\
B & Diff. & 4 & 1.02 & 0.005 & 0.003 & 0.582 & 0.000 & 0/5 & 0.856 & 0.925 \\
B & Diff. & 16 & 0.97 & 0.019 & 0.015 & 0.850 & 0.053 & 5/5 & 0.309 & 0.693 \\
B & Diff. & 64 & 1.08 & 0.021 & 0.008 & 0.913 & 0.261 & 3/5 & 0.179 & 0.614 \\
B & Diff. & 256 & 1.15 & 0.038 & 0.009 & 0.924 & 0.365 & 2/6 & 0.142 & 0.564 \\
 
\bottomrule
\end{tabular}
\end{table}

\begin{figure}[!htbp]
\centering
\includegraphics[width=\linewidth]{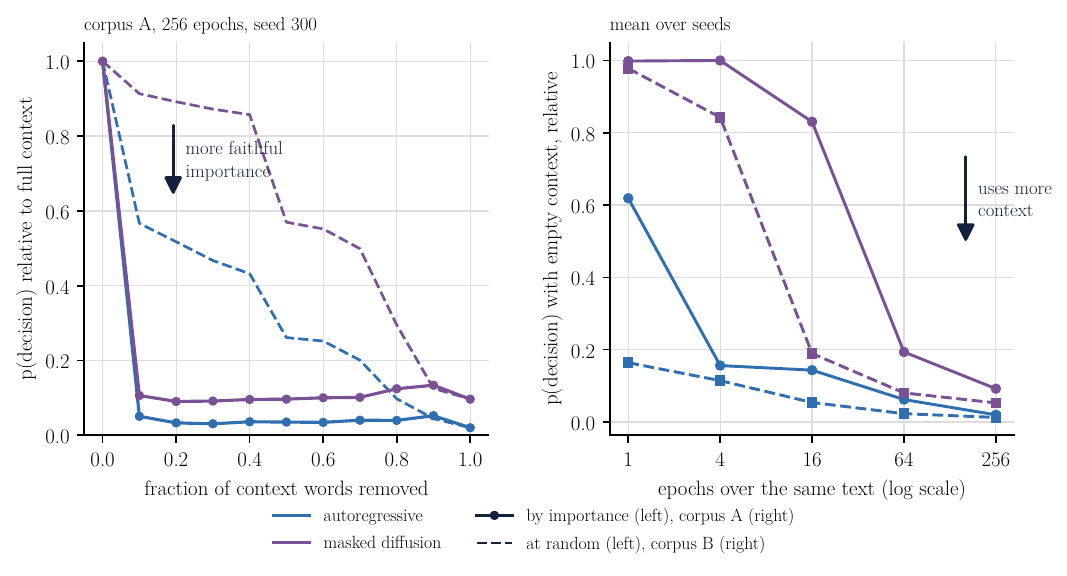}
\caption{Left, deletion curves of the first decisive seed of corpus A at 256 epochs, context words removed by importance (solid) and at random (dashed), mean over 300 test passages. Right, probability of the decision with the whole context removed, relative to the full context, across the budget grid, mean over seeds, corpus A (solid) and corpus B (dashed).}
\label{fig:deletion}
\end{figure}

The deletion curves show how much each prediction depends on its context. Removing the context words ranked as most important drops the probability of the decision much faster than removing words at random, with an area under the curve of 0.080 against 0.328 for the autoregressive mode at 256 epochs on corpus A and 0.149 against 0.605 for diffusion, so the ranking finds the words the decision rests on (\Cref{fig:deletion}). With the whole context removed, the remaining relative probability measures how much the model predicts from its own bias. For diffusion on corpus A it stays between 0.83 and 1.0 up to 16 epochs, so the early diffusion model predicts nearly the same token whatever the context, and it falls to 0.19 at 64 epochs and 0.09 at 256 as its likelihood improves.

\subsection{How much text the model needs}\label{sec:datacurve}

On corpus B the floor is just as close. The 6-gram floor on the same validation windows reaches 1.066 bits per byte, the value of the best autoregressive point to the third decimal, and is better than the best diffusion point (\Cref{fig:epochsB}). The tie holds on the same windows only, since on running text the floor reaches 1.038 and is ahead. Whether this reflects the method or the amount of text can be measured, since more text is the one resource a reader with a specialist corpus can sometimes add. A preregistered data curve trains the autoregressive mode, at 16 and 64 epochs with 3 seeds and the better budget on validation, and the n-gram on nested fractions of the training groups of corpus B, from one eighth to all of them, and scores both and their mixture on the 300 external laws, on the same windows. The Transformer was to count as worth its cost at a fraction where the lower limit of its gain over the floor passed $\Delta_{\min}$.

\begin{figure}[!htbp]
\centering
\includegraphics[width=0.7\linewidth]{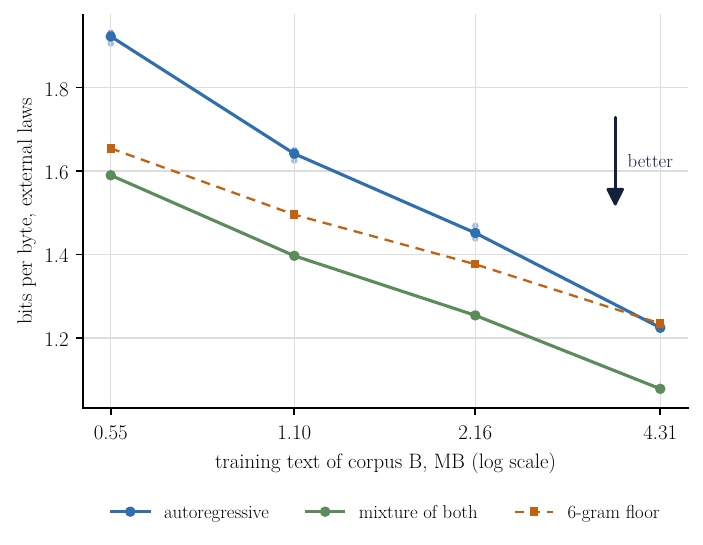}
\caption{Bits per byte on the external laws of 2025 and 2026 against the training text of corpus B, nested fractions of the training groups. Autoregressive Transformer (points are seeds), 6-gram floor and their mixture, with the mixing weight fitted on validation per seed.}
\label{fig:datacurve}
\end{figure}

No fraction met that criterion, but the curve shows where the answer lies (\Cref{fig:datacurve}). With 0.55\,MB of training text the Transformer is worse than the floor by 0.268 bits per byte, $[0.232, 0.305]$. Each doubling closes the gap, to 0.146 at 1.10\,MB and 0.075 at 2.16\,MB, and with the whole 4.31\,MB the Transformer is ahead by 0.010, $[-0.006, 0.026]$, a tie on the same windows. The grid ends there, so where the gain would pass $\Delta_{\min}$ is not measured. The curve fixes the number of epochs, so the compute grows with the text, and the hyperparameters come from the search on the whole corpus, so it measures more text and more compute together, not text alone. The mixture of the two beats the floor at every size, by 0.065 at 0.55\,MB and 0.157 at 4.31\,MB, and the weight it gives the Transformer rises from 0.30 to 0.57. As on corpus A, a reader with a few megabytes of text gets the most from keeping the n-gram and mixing.

\subsection{Usefulness on legal tasks}\label{sec:taskresults}

\begin{figure}[!htbp]
\centering
\includegraphics[width=\linewidth]{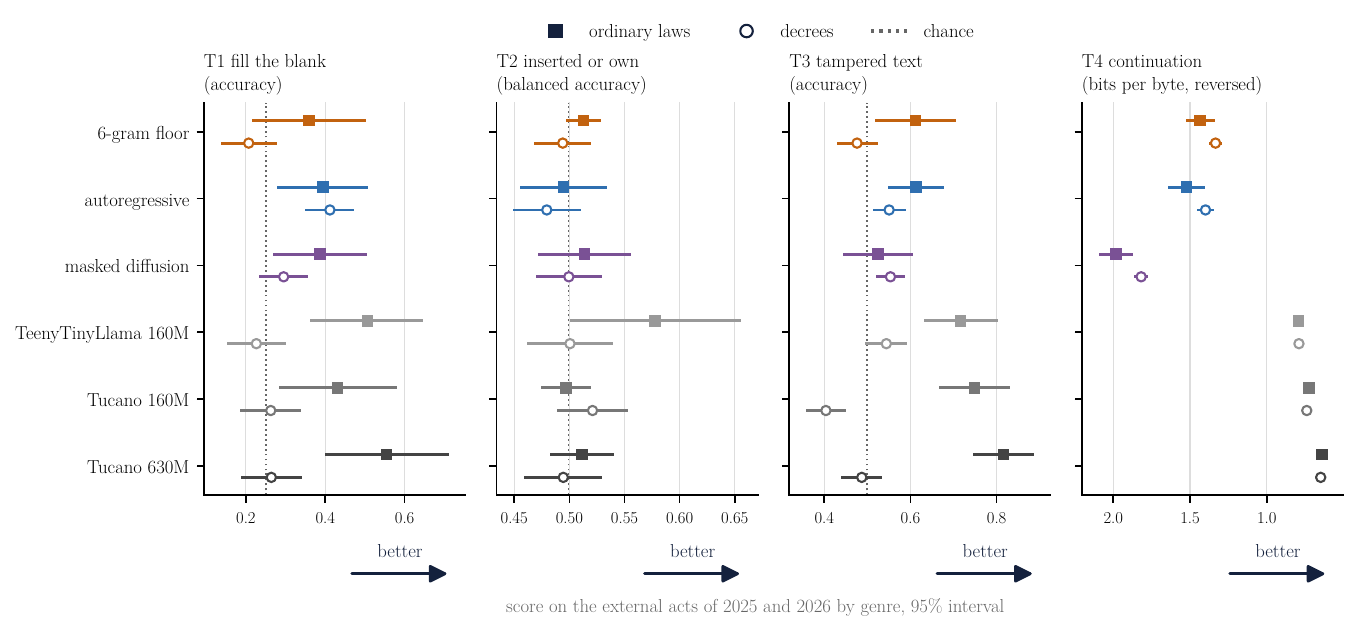}
\caption{Score on each task by genre of the act, ordinary laws (filled squares) and decrees (open circles), with the 95\% interval and chance as a dotted line. The T4 axis is reversed so that right is better in every panel. In decrees the floor and the public yardsticks stay near or below chance in T1 and T3.}
\label{fig:tasks}
\end{figure}

Of the three confirmatory tasks, the model beats the floor in one, loses in one and does not beat it in the third, and the descriptive task favours it (\Cref{tab:tasks}, \Cref{fig:tasks}). Detecting tampered text, the autoregressive mode beats the floor, with a difference of 0.062, $[0.014, 0.110]$, in favour in 202 of the 356 groups where the two differ, Holm-adjusted $p = 0.025$. Masked diffusion has an interval above zero as well, 0.053, $[0.015, 0.091]$, but its sign test does not reject, Holm-adjusted $p = 0.13$, so by the preregistered rule it does not beat the floor. Filling in the blank, descriptive by design, the autoregressive mode chooses the right number or cross-reference 0.399 of the time against 0.241 for the floor, a difference of 0.158, $[0.090, 0.226]$, and diffusion, the mode built for this task, gains 0.069, $[0.011, 0.128]$. Completing a provision, both modes are worse than the floor, the autoregressive mode by 0.070 bits per byte, $[0.056, 0.085]$, and the diffusion bound by 0.497, which by the reading of \Cref{sec:evaluation} says nothing about the true likelihood of diffusion. Telling an inserted provision from one of the law's own, no mode beats the floor, the autoregressive mode is below it, $[-0.064, -0.005]$, and diffusion ties it, $0.0004$, $[-0.024, 0.025]$. Removing the largest group changes none of these readings.

\begin{table}[!htbp]
\centering
\caption{Scores on the four tasks over the external acts of 2025 and 2026, for all acts and by genre, mean of the per-group values with a 95\% wild cluster bootstrap interval. T1 and T3 accuracy, T2 balanced accuracy, T4 bits per byte (lower is better). Complementary laws enter the total only, with 7 to 17 groups per task. The project's model is at the best validation point of each mode, 5 seeds for the autoregressive mode and 6 for diffusion, whose T4 value is an upper bound. T2 is exploratory.}
\label{tab:tasks}
\scriptsize
\begin{tabular}{lcccc}
\toprule
Scorer & T1 Filling in the blank & T2 Inserted or own & T3 Tampered text & T4 Continuation \\
\midrule
\emph{All acts} & \emph{130 groups} & \emph{643 groups} & \emph{469 groups} & \emph{615 groups} \\
6-gram floor & 0.241 [0.179, 0.301] & 0.502 [0.488, 0.515] & 0.497 [0.456, 0.537] & 1.355 [1.324, 1.385] \\
autoregressive & 0.399 [0.349, 0.450] & 0.467 [0.442, 0.492] & 0.559 [0.528, 0.589] & 1.425 [1.383, 1.468] \\
masked diffusion & 0.310 [0.258, 0.361] & 0.502 [0.476, 0.529] & 0.550 [0.523, 0.578] & 1.852 [1.816, 1.888] \\
TeenyTinyLlama 160M & 0.269 [0.205, 0.331] & 0.541 [0.495, 0.588] & 0.567 [0.527, 0.607] & 0.791 [0.779, 0.804] \\
Tucano 160M & 0.294 [0.227, 0.358] & 0.510 [0.491, 0.529] & 0.458 [0.419, 0.498] & 0.737 [0.726, 0.748] \\
Tucano 630M & 0.322 [0.253, 0.390] & 0.498 [0.479, 0.518] & 0.538 [0.498, 0.579] & 0.647 [0.635, 0.660] \\
\midrule
\emph{Ordinary laws} & \emph{23 groups} & \emph{148 groups} & \emph{70 groups} & \emph{133 groups} \\
6-gram floor & 0.359 [0.219, 0.499] & 0.513 [0.498, 0.528] & 0.611 [0.520, 0.701] & 1.435 [1.349, 1.520] \\
autoregressive & 0.395 [0.283, 0.505] & 0.495 [0.457, 0.533] & 0.613 [0.552, 0.674] & 1.523 [1.409, 1.636] \\
masked diffusion & 0.386 [0.271, 0.502] & 0.514 [0.473, 0.555] & 0.525 [0.447, 0.602] & 1.983 [1.881, 2.086] \\
TeenyTinyLlama 160M & 0.506 [0.366, 0.643] & 0.578 [0.502, 0.655] & 0.716 [0.634, 0.799] & 0.793 [0.767, 0.819] \\
Tucano 160M & 0.431 [0.287, 0.576] & 0.497 [0.476, 0.518] & 0.749 [0.669, 0.829] & 0.724 [0.700, 0.748] \\
Tucano 630M & 0.554 [0.402, 0.707] & 0.511 [0.484, 0.539] & 0.816 [0.748, 0.883] & 0.642 [0.620, 0.664] \\
\midrule
\emph{Decrees} & \emph{104 groups} & \emph{492 groups} & \emph{396 groups} & \emph{480 groups} \\
6-gram floor & 0.207 [0.140, 0.274] & 0.494 [0.469, 0.518] & 0.476 [0.432, 0.521] & 1.334 [1.304, 1.364] \\
autoregressive & 0.412 [0.353, 0.469] & 0.479 [0.450, 0.509] & 0.551 [0.516, 0.586] & 1.399 [1.356, 1.443] \\
masked diffusion & 0.295 [0.236, 0.352] & 0.500 [0.471, 0.528] & 0.554 [0.524, 0.583] & 1.818 [1.783, 1.854] \\
TeenyTinyLlama 160M & 0.226 [0.156, 0.297] & 0.501 [0.463, 0.538] & 0.544 [0.498, 0.588] & 0.791 [0.776, 0.806] \\
Tucano 160M & 0.263 [0.189, 0.336] & 0.521 [0.490, 0.552] & 0.404 [0.361, 0.447] & 0.740 [0.727, 0.752] \\
Tucano 630M & 0.264 [0.191, 0.338] & 0.494 [0.460, 0.529] & 0.487 [0.443, 0.531] & 0.649 [0.634, 0.663] \\
 
\bottomrule
\end{tabular}
\end{table}

One reading of T2 departs from the preregistration and is recorded as a deviation made after the measurement. The preregistered rule applies the sign test to the contribution of each group to the balanced accuracy. On the external set, 547 of the 643 groups hold provisions of one class only, and there the difference between two scorers measures only their accuracy on the majority class. The test then rejects with about 85\% of the groups in favour of any scorer, whether the balanced accuracy goes up or down, so T2 is read by the interval of the balanced accuracy instead. The output of the original rule is kept in the repository.

Seven in ten groups of the external set are decrees, a genre absent from training, and the reading changes with the genre (\Cref{tab:tasks}, \Cref{fig:tasks}). On ordinary laws the advantage in tampered text disappears, 0.613 for the autoregressive mode against 0.611 for the floor over 70 groups, the continuation keeps its order, 1.523 against 1.435, and the public yardsticks, pretrained on much more text, are ahead of the home-built model by point estimate in every task but T2, with 0.716 to 0.816 in tampered text, 0.43 to 0.55 in filling the blank and 0.64 to 0.79 bits per byte in continuation. In decrees the autoregressive mode leads T1 and T3, at 0.412 and 0.551, but there the floor and the yardsticks stay near or below chance, 0.207 to 0.264 in T1 against a chance of 0.25 and 0.404 to 0.544 in T3 against 0.5. The lead in T1 depends on weighting each group equally. Weighting each item instead, the autoregressive mode and Tucano 630M tie over all acts, at 0.401 each. T1 also mixes two classes of very different size, 308 cross-references and 29 numbers, all of the latter in laws, and on numbers the yardsticks lead, 0.52 to 0.76 against 0.35 for the autoregressive mode (\Cref{tab:diagnostic}), so no claim is made here for deleted numbers. In T2 every scorer, Tucano 630M included, stays between 0.46 and 0.55 overall, so the task does not separate the scorers.

Scorers near or below chance point to items that a likelihood score gets systematically wrong, so the stored choices were examined for three causes, with no new run (\Cref{tab:diagnostic}). The first is surface frequency, the competition among surface forms that \citet{holtzman2021surface} correct with a conditional pointwise mutual information score, not applied here. The value that tells the candidates apart was counted in the training text of corpus B, which the home-built models and the floor learned from and the yardsticks never saw, so the measure is asymmetric by construction. The home-built models and the floor pick the most frequent of the four options of T1 in 0.48 to 0.69 of the items, against 0.24 to 0.31 for the yardsticks and 0.33 for the right answer being the most frequent. Their accuracy follows that prior. When the right answer is the most frequent option the autoregressive mode is right in 0.564 of the items and otherwise in 0.321, diffusion in 0.673 and 0.147 and the floor in 0.450 and 0.164, while the yardsticks barely change, at 0.35 to 0.39 when it is not and 0.37 to 0.41 when it is. In T3 the preference for the more frequent of the two versions separates only the home-built modes, 0.68 and 0.70, from the floor and the yardsticks, 0.42 to 0.59, against a base rate of 0.55.

The second cause is the weight of small groups. Each group weighs the same, and 67 of the 104 decree groups of T1 and 265 of the 396 of T3 hold a single item, nearly all of them cross-references. In those single-item decrees of T1 the yardsticks score 0.18 to 0.19, at chance by a one-sided binomial test over the 67 items ($p$ from 0.11 to 0.18), and only the floor, at 0.149, is below it ($p = 0.034$). In those of T3 the floor and the two Tucano models are below chance, 0.29 to 0.43 over 265 items ($p \le 0.02$), and TeenyTinyLlama, at 0.475, is not. In the larger decree groups the scorers are at or above chance in T1, the yardsticks and the home-built modes at 0.34 to 0.42, and all of them are above chance in T3, from 0.58 to 0.70.

The third cause concerns what the home-built model knows. A decree regulates a law, the law whose article it cites may be in the training text of corpus B, and the 13-token filter removes an item only when its wording is copied. Classifying the cross-references of decrees in T1 by the act named right after the gap, 38 point to a law of the training partition, 53 to another law and 5 to the decree itself, and 135 name no act there. On the 38 the autoregressive mode is right in 0.511 of the items, against 0.374 on the 53, while the floor, diffusion and the yardsticks score from 0.026 to 0.184 on them. Part of the advantage of the autoregressive mode in decrees therefore comes from knowing the laws they regulate, not from generalising to a new genre, a reading that rests on a heuristic classification and few items. Why the other scorers prefer the distractors on those items was not established, so the validity of the T1 and T3 items on decrees is not established either, and the advantages of the home-built model that come from the decrees are reported as such, not as evidence of usefulness.

\begin{table}[!htbp]
\centering
\caption{Diagnostic of T1 and T3 on the external acts from the stored choices, item accuracy or rate, mean over seeds. Laws include the complementary laws. Most frequent refers to the count, in the training text of corpus B, of the value that tells the candidates apart, the number or the cited article in T1, and the term, number or cited article in T3, where shuffled subdivisions have no such value. Single-item groups are decree groups with one item, which weigh as much as any other group in \Cref{tab:tasks}.}
\label{tab:diagnostic}
\scriptsize
\begin{tabular}{lcccccccc}
\toprule
 & \multicolumn{3}{c}{T1 accuracy by class} & \multicolumn{2}{c}{Picks the most frequent} & \multicolumn{3}{c}{Accuracy in decrees} \\
\cmidrule(lr){2-4}\cmidrule(lr){5-6}\cmidrule(lr){7-9}
Scorer & \makecell{Numbers,\\ laws (29)} & \makecell{Cross-ref.,\\ laws (77)} & \makecell{Cross-ref.,\\ decrees (231)} & \makecell{T1 of 4\\ options} & \makecell{T3 of 2\\ versions} & \makecell{T1, single-\\item groups} & \makecell{T3, single-\\item groups} & \makecell{T3, other\\ groups} \\
\midrule
6-gram floor & 0.379 & 0.286 & 0.234 & 0.487 & 0.557 & 0.149 & 0.377 & 0.695 \\
autoregressive & 0.345 & 0.395 & 0.410 & 0.476 & 0.676 & 0.385 & 0.482 & 0.659 \\
masked diffusion & 0.425 & 0.301 & 0.314 & 0.689 & 0.699 & 0.234 & 0.526 & 0.580 \\
TeenyTinyLlama 160M & 0.517 & 0.481 & 0.294 & 0.309 & 0.586 & 0.179 & 0.475 & 0.639 \\
Tucano 160M & 0.621 & 0.390 & 0.329 & 0.243 & 0.422 & 0.194 & 0.294 & 0.682 \\
Tucano 630M & 0.759 & 0.455 & 0.338 & 0.267 & 0.577 & 0.194 & 0.434 & 0.623 \\
 
\bottomrule
\end{tabular}
\end{table}

\begin{figure}[!htbp]
\centering
\includegraphics[width=\linewidth]{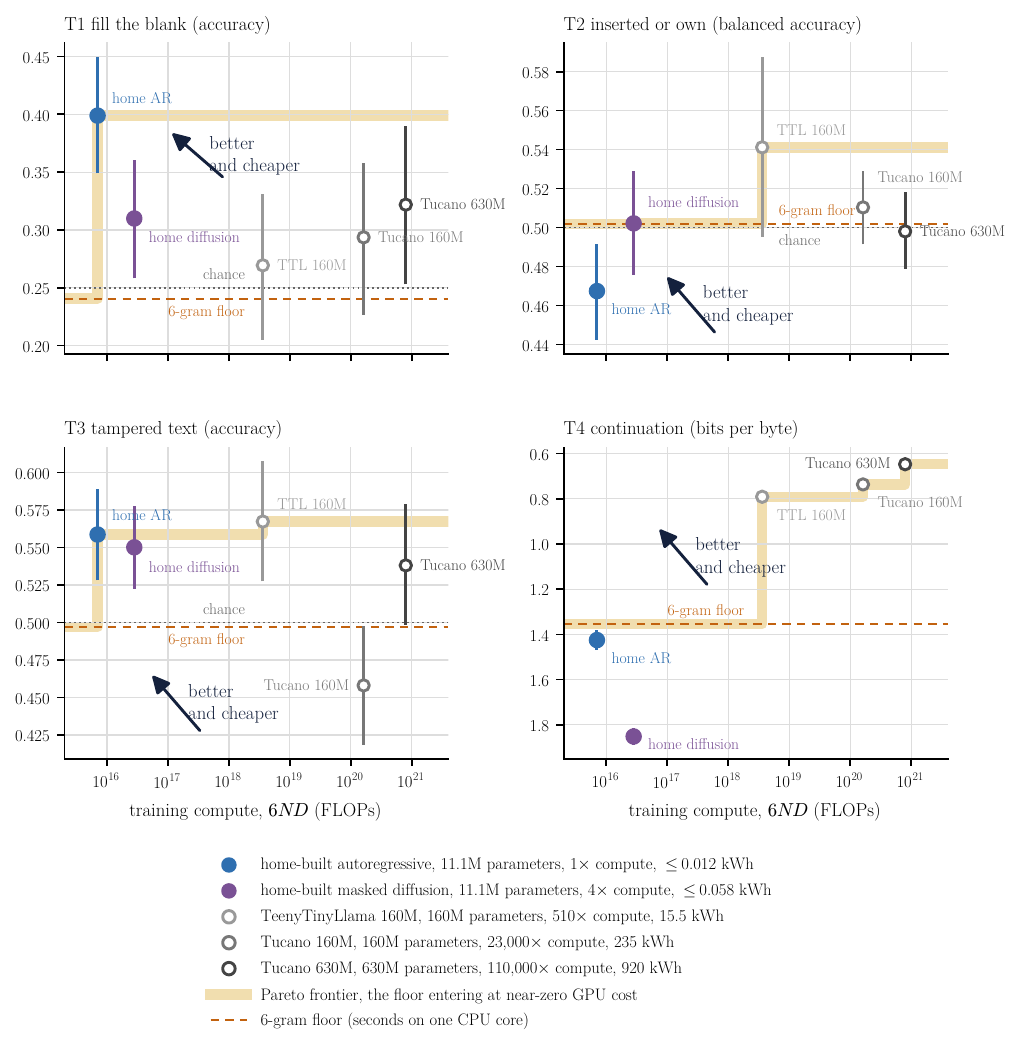}
\caption{Score on each task of the external set against training compute, $6ND$, on a logarithmic scale. Filled points are the home-built model, open points the public yardsticks, with 95\% intervals. The shaded steps are the Pareto frontier, where no cheaper scorer does better, with the 6-gram floor entering at the left edge because it trains in seconds on one CPU core. The T4 axis is reversed so that better is up in every panel. The legend gives parameters, compute relative to the home-built autoregressive model to two significant figures and training energy, an upper bound at the 500\,W cap for the home-built model and the values reported by the authors for the yardsticks. Scores are over all acts, so the T1 and T3 panels carry the decrees, where the floor and the yardsticks stay near or below chance (\Cref{fig:tasks}).}
\label{fig:frontier}
\end{figure}

Scores alone hide what each of them cost. \Cref{fig:frontier} places every scorer by the compute of its training run, approximated by $6ND$ with $N$ parameters and $D$ token positions seen in training, the tokens, energy and parameters of the yardsticks taken from their papers \citep{correa2024teenytiny,correa2024tucano}. A scorer is on the Pareto frontier when no cheaper one scores higher, and the floor enters the frontier at the left edge, since it trains in seconds on one CPU core. The yardsticks cost about $5 \times 10^2$ to $10^5$ times the compute of the home-built autoregressive model and about $10^3$ to $7 \times 10^4$ times its energy, orders of magnitude rather than exact ratios. The two energies are not measured the same way, an upper bound for the GPU at the power cap here and the values reported by the authors for the yardsticks, so that ratio is an order of magnitude. The five models are ordered the same way by parameters, compute, energy and wall time, so the frontier does not depend on which of these measures the cost. Over all acts, in T1 the home-built autoregressive model is the whole frontier after the floor, ahead of every yardstick, although its interval touches that of Tucano 630M and the lead becomes a tie with it when items rather than groups are weighted, and in T3 it shares the frontier with TeenyTinyLlama 160M, whose point estimate is higher by 0.008 at about 500 times the compute, with overlapping intervals. Both readings are carried by the decrees, where the floor and the yardsticks are near or below chance. On ordinary laws alone every yardstick is ahead of the home-built model by point estimate in T1, T3 and T4, and on a frontier drawn for the laws the home-built model would stay only as the cheapest neural point, by 0.036 over the floor in T1 and 0.002 in T3. In T4 both home-built modes fall off the frontier, below the cheaper floor, and the frontier is made of the floor and the three yardsticks. In T2 the frontier moves by differences inside the noise, and masked diffusion sits on it by 0.0004 over the floor. The frontier uses point estimates, so a scorer on it whose interval overlaps a neighbour's is tied with it, and the reading carries the asymmetries of scale and domain set out in \Cref{sec:limitations}.

\subsection{A typed decider from the same network}\label{sec:deciderresults}

The network turns into a typed decider with a short module, but in this experiment the decider is not better than the lexical floor (\Cref{tab:decider}). In T2 the deciders on both modes reach 0.499 and 0.488 against 0.544 for TF-IDF, and in T3 they stay at chance, 0.501 and 0.514 against 0.573. The cross-validation on training items already showed it, 0.49 to 0.51 at every learning rate in T3, and the multilingual Laya, an encoder of 322M parameters fine-tuned by the same protocol, fails the same way, 0.485 and 0.500, so the failure cannot be attributed to the home-built network alone.

Sanity checks run after the measurement, on the first seed of each mode, separate a broken pipeline from a failed training (\texttt{scripts/sanidade\_decisor.py}). Fine-tuned on 32 training texts, half of each class, for 300 full-batch steps, the decider fits them exactly at every learning rate of the grid and at $10^{-3}$, in both modes and both tasks, with a final loss below $10^{-3}$, so the module can learn. Under the protocol, the outcome differs by task. In T2 the training loss falls and the decider reaches a balanced accuracy of 0.79 and 0.74 on its own training items, so it learns them and does not carry what it learned to the external acts. In T3 the training loss stays between 0.69 and 0.76 from the first to the last of the 315 steps, around $\ln 2 \approx 0.693$, the balanced accuracy on the training items is 0.51 and 0.50, and the predictions collapse towards one class, with the share of texts called tampered on the external set ranging from 0 to 1 across seeds. That is the signature of a constant predictor, compatible with training that did not succeed under three epochs at a constant rate of at most $10^{-4}$, and its cause is not attributed here.

Calibration says the same thing from another side. In T3 the fitted temperature reaches the end of its search range, $e^4$, which flattens every probability to about 0.5, and the calibration error falls near zero. A decider that knows nothing and says so is well calibrated, so a low calibration error is meaningful only next to the accuracy. In T2 the temperature lowered the error in 7 of the 11 runs. The selective answer, with its threshold fixed out of fold on training for 95\% accuracy, covers at most 2.3\% of the external items on average, and the covered items reach 0.51 to 0.89, so a threshold learned on the training laws does not carry over to acts of another period and genre.

\begin{table}[!htbp]
\centering
\caption{Typed deciders on the external acts of 2025 and 2026. T2 balanced accuracy and T3 accuracy, mean of the per-group values with a 95\% interval. Selective answer at a 95\% target, mean coverage over seeds and, in parentheses, the accuracy of the covered items.}
\label{tab:decider}
\scriptsize
\begin{tabular}{lcccc}
\toprule
 & \multicolumn{2}{c}{Score} & \multicolumn{2}{c}{Selective coverage} \\
\cmidrule(lr){2-3}\cmidrule(lr){4-5}
Decider & T2 & T3 & T2 & T3 \\
\midrule
TF-IDF with logistic regression (floor) & 0.544 [0.487, 0.600] & 0.573 [0.553, 0.593] &  &  \\
autoregressive checkpoint & 0.499 [0.459, 0.539] & 0.501 [0.500, 0.503] & 0.90\% (0.89) & 2.29\% (0.51) \\
masked diffusion checkpoint & 0.488 [0.441, 0.536] & 0.514 [0.511, 0.517] & 0.05\% (0.67) & 0.59\% (0.54) \\
Laya multilingual, fine-tuned & 0.485 [0.446, 0.524] & 0.500 [0.494, 0.506] & 0.00\% & 0.00\% \\
Laya multilingual, no fine-tuning & 0.515 [0.494, 0.535] & 0.499 [0.492, 0.505] &  &  \\
 
\bottomrule
\end{tabular}
\end{table}

\subsection{Synthesis}

The question of the tutorial has a measured answer. A language model of 11 to 12M parameters, built from nothing on a home GPU in minutes per run, reaches the level of a well-smoothed n-gram scored on the same windows, ties it on the legal corpus and passes it once the two are mixed. On acts it has never seen, its autoregressive mode does not complete provisions better than counting does. It beats the n-gram at detecting tampered text, and fills deleted cross-references more often than the n-gram and public models of 160M and 630M parameters, only in decrees, a genre absent from training, where those scorers stay near or below chance and where the validity of the items is not established. That advantage rests in part on a frequency prior learned from its training text and on knowing the laws that the decrees regulate, and in filling the blank it holds only when each group weighs the same. On ordinary laws the public models are ahead. None of this is a reason not to brew one. It is the map a reader needs before brewing, since the same scripts that produced these numbers will say, for a new corpus, where the home-built model earns its place. More text, together with the compute that comes with it, is the clearest lever measured here, and the mixture with the n-gram is the cheapest gain. Between the two modes, the choice is one of budget. At their best points, on test, the diffusion bound is above the autoregressive likelihood by less than the minimum effect. The autoregressive mode learns faster and memorises its training text when read too many times, which shows in its likelihood, in its calibration and in its samples, while masked diffusion tolerates repetition and reaches its best point later, an ordering at 256 epochs that holds on the test partition and across the factors varied here. The blind reading, the semantic judge and the reproduction on the T4 are designed, with their criteria frozen, and enter a later version.

\section{Checks that keep defects out of the numbers}\label{sec:checks}

A pipeline written by hand will contain defects, and many of them do not stop the code from running. They change a number. \Cref{tab:checks} lists the checks that detected a defect in this project, what each one detected and the fix. Most of these defects are known in the literature, and a reader who builds their own pipeline is likely to meet several of them, which is why each check is part of the package and runs again after every change.

\begin{table}[!htbp]
\centering
\caption{Checks that detected a defect during construction, with what was measured and the fix adopted. Numbers come from the tests and scripts of the repository.}
\label{tab:checks}
\scriptsize
\begin{tabularx}{\linewidth}{@{}>{\raggedright\arraybackslash}p{2.6cm}>{\raggedright\arraybackslash}X>{\raggedright\arraybackslash}p{3.6cm}@{}}
\toprule
Check & What it detected & Fix \\
\midrule
Length test with exact conditionals & If diffusion receives the document length without paying for it, it gains the entropy of the length, 0.856 nat (1.235 bits) per document in the test, matched within $10^{-9}$ & Predict \EOS{} and padding in both modes \\
Weight of the loss & Weight $1/t$ unbounded near $t = 0$, infinite variance without a cut-off, and the cut-off truncates the bound & Number-of-masks form, weight at most $L$ (\Cref{prop:nmasks,prop:variance}) \\
Logit gradient in fp16 & With one masked token per sequence and scale $2^{16}$, the logit gradient exceeds the fp16 maximum of 65,504 for the mean over masked tokens with weight $1/t$ at batch 2, 4 and 64 & Number-of-masks form, peak bounded by scale over batch, 16,384 at batch 4 \\
Precision sentinel & Median per-tensor error of 1.38 and 1.45, maximum 7.11, against a margin of 0.05, caused by dropout in the sentinel itself & Both passes of the sentinel in eval mode, with a test \\
Range of the sampling noise & In fp16 the uniform draw has 2,048 values and the largest Gumbel noise is 7.62, against 16.64 in fp32 and 36.74 in fp64 & Categorical draws in float64 \citep{zheng2024timeagnostic} \\
Enumeration of all sequences & With a free \MASK{} logit the distribution sums to 0.5906 instead of 1, and with a target visible to its own prediction to 1.2837 & \MASK{} logit at $-\infty$ \\
Equality of two BPE implementations & In 3,463 of 3,836 merges (90\%) more than one pair tied at the maximum count, the first at merge 56 & Written tie-breaking rule, merge-by-merge test \\
Vocabulary absent from held-out text & With $V = 8192$, 12.0\% of the learned tokens never occur in validation, against 3.5\% with $V = 4096$ & Largest $V$ with at most 10\% absent \\
Unicode normalisation & Under NFD the pre-tokenisation pattern splits a combining mark from its letter & NFC once, before everything \\
Names across blocks & In a sample of 4 works, 46\% to 97\% of the names of a held-out block already occur in training & Bits per byte with and without name spans \\
Blind reading design & Model samples start at a document start and human excerpts in the middle of one, so the first words reveal the origin & Crop of 60 words from a random position in both \\
Copy from training & Longest copied span, median of 4 words at 16 epochs and 144 to 150 at 256 epochs without dropout, maximum 180 & Dropout, copy filter, discard rate reported \\
Cost estimate & $6N$ per token ignores attention and the output layer, 73\% of the cost in a 4-layer model with $d = 256$ and $V = 8192$ & Full count, throughput measured \\
Calibration on validation predictions & Taken in document order, the 50,000 predictions came from the first works only, the two halves of the selective check had accuracies of 0.178 and 0.084, and the padding after \EOS{} made up 24\% to 90\% of the accepted high-confidence predictions & Windows in a fixed random order, padding after \EOS{} left out, with a test \\
Monte Carlo error of the bound & Required below $\Delta_{\min}/4$ by the budget check and not computed before the pilot. Measured after the final runs, 0.034 bits per byte for a single evaluation with $K = 1$ & Reported, $K \ge 4$ for single runs (\Cref{sec:sensitivity}) \\
Decider sanity & A small batch is fitted exactly, but under the protocol the T3 training loss stays near $\ln 2$ and the predictions collapse to one class & Reported as a training failure of unattributed cause, kept out of the summary claims (\Cref{sec:deciderresults}) \\
Scorers below chance & In single-item decree groups, mostly cross-references, the floor falls below chance in T1, and the floor and both Tucano models in T3, while the home-built model does best where the decree cites a law of its training text & Results by genre and per item, validity of the decree items not established (\Cref{sec:taskresults}) \\
\bottomrule
\end{tabularx}
\end{table}

\begin{figure}[!htbp]
\centering
\includegraphics[width=0.7\linewidth]{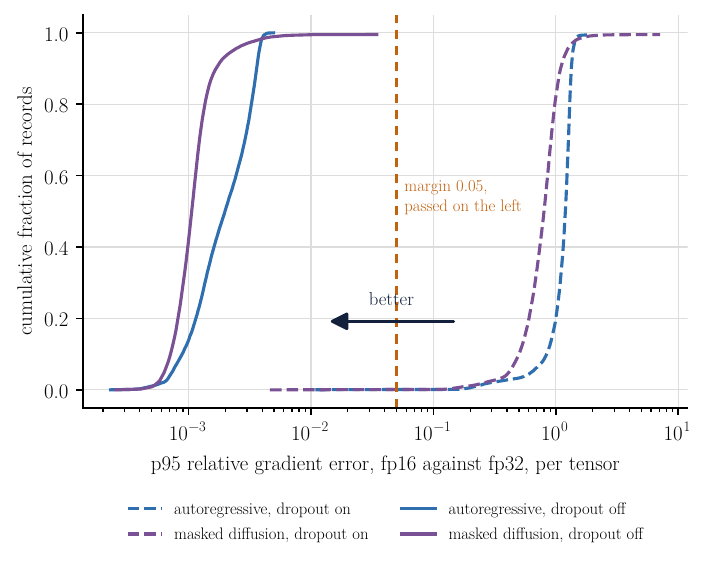}
\caption{Cumulative distribution of the per-tensor p95 relative error between fp16 and fp32 gradients over all sentinel records of the 12 final runs at 256 epochs. Dashed, the sentinel in training mode, where each pass draws its own dropout mask. Solid, the sentinel in eval mode. The vertical line is the preregistered margin of 0.05.}
\label{fig:sentinel}
\end{figure}

\paragraph{A check of the checking instrument.} The precision sentinel compares, tensor by tensor, the gradient in fp16 with the gradient in fp32 on fixed probe batches, from the start to the end of training. In the calibration runs the network had no dropout, and the sentinel passed with a worst per-tensor error of 0.0025 in the autoregressive mode and 0.0099 in diffusion. The hyperparameter search then chose dropout of 0.2 and 0.1, and in the first final runs the same sentinel reported a median per-tensor error of 1.38 and 1.45 and a maximum of 7.11. Every one of the twelve runs exceeded the margin of 0.05, which by the preregistered rule would have removed all of them from the comparison. The cause was in the sentinel. Both backward passes ran in training mode, each drew its own dropout mask, and the error measured the distance between two masks rather than two precisions. With both passes in eval mode, the median error fell to 0.0020 and 0.0046 and the maximum to 0.0049 and 0.0346 (\Cref{fig:sentinel}). The final runs were repeated with the corrected sentinel, and the estimate of $\Delta$ moved from $-1.028$ to $-1.026$. A unit test now fails if the sentinel runs with dropout on. An instrument validated in one configuration needs to be validated again when the configuration changes.

\paragraph{Paying for the length.} An autoregressive model pays for the end of a document through the probability of \EOS. A diffusion model that receives its input already cut to the document's length does not, and the comparison would then favour diffusion by up to $\log_2 L$ bits per document. The package test builds documents of random length with exact conditionals, so that the two modes differ only in how they handle length. The advantage of diffusion equals the entropy of the length distribution within $10^{-9}$ and vanishes when both modes predict the padding, at the price of 12.2\% of the training positions.

\paragraph{Recitation as a form of overfitting.} In the search, the autoregressive mode without dropout and with learning rate $10^{-3}$ reached 5.6 bits per byte on validation after 256 epochs, far worse than after 16. Its samples changed in the same direction. Over 16 samples of 256 tokens per run, the longest span copied from training had a median of 4 words at 16 epochs and of 144 to 150 words at 256, with a maximum of 180. With dropout 0.2 and learning rate $3 \times 10^{-4}$, the median at 256 epochs fell to 5 to 5.5 words, with isolated samples of 44 and 95 copied words. A fluent sample can therefore be a copy, which is why samples are filtered by copy before any reading or semantic scoring.

\section{Adapting the tutorial to a new domain}\label{sec:adapting}

The steps below move the pipeline to a new corpus. Each step has a recipe in the cookbook of the repository.
\begin{enumerate}
\item Write the domain adapter that produces documents with an identifier, a group and the cleaned text, normalised to NFC.
\item Split by group into training, validation and test, and measure the size of each partition.
\item Train the tokenizer on the training partition and choose $V$ by the rule of absent tokens on the validation text.
\item Measure the padding fraction for candidate window lengths and fix $L$.
\item Run the formulation checks on CPU and measure the throughput of the machine.
\item Search the learning rate and dropout on the new corpus, freeze $\Delta_{\min}$, run the pilot and derive the number of seeds.
\item Train the budget grid in each mode and evaluate likelihood, the references, the tasks of the domain and the quality of the samples.
\end{enumerate}
The tasks of a new domain follow the same pattern as the legal ones. Collect an external set dated after the training text and after the public models in use, grouped by the same rule. Derive each task from the text alone, by deleting a span that the domain fixes, such as a number or a reference, by altering it with operations that keep its surface plausible, or by cutting it, so that no labelling is needed. Discard items that share a long span with training and items that do not fit the window. Count items and groups, compute the power of the sign test over groups, and freeze the design, the item hashes and the decision rule before scoring anything. Check that the test of each task measures its metric when the classes are unbalanced within groups, as T2 here shows it may not.

\section{Limitations}\label{sec:limitations}

The model has 12.3M parameters and the main corpus 0.62M training tokens, and nothing here transfers to larger scale without new measurement. The claims on corpus A hold for unseen chapters of works whose other chapters were in training, proper names cross the blocks, and the analysis that leaves a whole work out is secondary. For diffusion, the reported likelihood is a bound whose gap to the true value is not measured at this scale, so a result in which the autoregressive mode wins says nothing about the true likelihood of diffusion.

On corpus A the bootstrap failed its coverage check, so the result at 256 epochs rests on a randomization test over 6 seeds whose smallest attainable $p$ is 0.031. Every seed agreed in sign by a wide margin, but a test at its floor cannot express more evidence than that. One amendment was made after seeing data and is recorded as such, and the comparison of the best points of the two modes, by TOST, was added after the decisive analysis. The margins of the sentinel, the threshold of 10\% in the vocabulary rule, the coverage minimum and the agreement threshold of the reading have no source and are choices declared before the measurements they govern. The hyperparameter grid constrains the result at 256 epochs, as discussed in \Cref{sec:repetition}.

The levels N2 to N4, the semantic judge and the reproduction on the T4 are not part of this version, so nothing here says how coherent the samples are beyond the lexical level or how long the runs take on Colab. When measured, the levels will depend on a reading by three raters, two of them students who may lack training in nineteenth-century spelling, and the controls limit but do not remove the damage. The sensitivity analysis changed one factor at a time with 3 seeds on corpus A only. The calibration error at the fitted temperature is measured on the predictions used to fit it, one parameter on 50,000 predictions, and only the selective check is evaluated on held-out predictions. The generative perplexity uses a scorer that may have seen the prose of corpus A in pretraining. Corpus B validation and test hold 21 and 19 laws, and the group of the Civil Code holds 53\% of the test bytes, which is why usefulness is measured on the external acts. Seven in ten of those acts are decrees, a genre absent from training, so the task results mix the transfer to a new period with the transfer to a new genre, and the per-genre readings rest on fewer groups. In the decrees the floor and the yardsticks stay at chance in T1 and partly below it in T3, concentrated in single-item groups of cross-references, and the validity of those items is not established. The home-built model also leans on how often a candidate occurs in its training text and on knowing the laws that the decrees regulate, a score corrected for surface frequency was not tried, and the classification of the cited acts is a heuristic over few items. The audit of the class rule over 72 sampled summaries, which governs the selection of laws in corpus B, is pending. In T2 the preregistered sign test did not measure the balanced accuracy, and the reading by interval was chosen after the measurement. The decider used the default settings of its optimiser, a constant learning rate and three epochs for every model, and a protocol tuned per model might change its result, for the home-built network and for Laya alike. The external acts postdate the public yardsticks, but their wording repeats earlier laws that those models may have seen. The peak throughput behind the utilisation figures comes from secondary sources and was not checked against the vendor's documentation, so those figures are approximate.

The comparison with the public yardsticks is not between equals, on two axes that pull in opposite directions. In scale, the yardsticks are 14 to 57 times larger and saw far more text, about 3.75B tokens for TeenyTinyLlama 160M (8,192 tokens per batch over 458,000 steps, \citealp[Table 3]{correa2024teenytiny}) and 169B and 211B tokens for Tucano 160M and 630M \citep[Table 5]{correa2024tucano}, against 105M token positions for the home-built autoregressive model at its best budget on corpus B. Their training compute, approximated by $6ND$, is therefore about $5 \times 10^2$ to $10^5$ times larger. The token counts come from different tokenizers and give orders of magnitude, not exact ratios. In domain, the home-built model trained on the laws of corpus B, while the yardsticks were used as released, without adaptation, although the training corpus of Tucano holds more than 4B tokens from the legal domain \citep{correa2024tucano}. The task results therefore do not separate the effect of building from scratch from the effect of training in the domain. Where the home-built model falls short, in the continuation task and on the laws in T3, the gap may come from the amount of training, from the architecture and its specification (six layers of width 384, a window of 256 tokens, a vocabulary of 1,024), or from both, and none of these causes was tested here. The control that would separate them, a public model of similar size adapted on the same training text with the same budget, was not run, and a Portuguese legal model such as RoBERTaLexPT \citep{garcia2024robertalexpt} is a candidate for it.

\section{Reproducibility}\label{sec:repro}

Every number in this tutorial comes from a script of the repository, and every run records its configuration, the hashes of the corpus and of the tokenizer, the versions of Python, PyTorch and CUDA, the GPU, the driver and the power cap. The cookbook gives one recipe per stage, from the environment to the blind reading, and the preregistrations with their hashes are versioned. The byte-level control uses \texttt{scripts/treinar.py} with a tokenizer folder whose metadata declare the byte implementation, the window and the epochs of \Cref{tab:bytes}. The per-run calibration and deletion come from \texttt{scripts/analisar\_corrida.py}, the scores on the test and training partitions from \texttt{scripts/avaliar\_particao.py}, the sensitivity analysis from \texttt{scripts/sensibilidade.py}, the entropy diagnostic from \texttt{scripts/entropia\_genppl.py}, the comparison of the best points from \texttt{scripts/melhor\_contra\_melhor.py}, the frequency diagnostic of the tasks from \texttt{scripts/diagnostico\_frequencia.py} and the decider sanity checks from \texttt{scripts/sanidade\_decisor.py}. The figures and tables are produced by \texttt{scripts/figuras\_artigo.py} and the bibliography by \texttt{scripts/bib\_artigo.py}, which accepts only sources downloaded and checked by the repository. The repository is at \url{https://github.com/pcbrom/homebrewed-slms}. The two notebooks open in Colab from \url{https://colab.research.google.com/github/pcbrom/homebrewed-slms/blob/main/notebooks/slm_autorregressivo.ipynb} and \url{https://colab.research.google.com/github/pcbrom/homebrewed-slms/blob/main/notebooks/slm_difusao.ipynb}.

\section*{Data and code availability}
Code, notebooks, preregistrations, cleaned corpora with manifests and per-run results are released under CC BY-NC-SA 4.0, except the corpus texts, which keep their own status as public domain works and official acts. CC BY-NC-SA 4.0 is not a software licence. It allows use, adaptation and redistribution for non-commercial research and teaching, with attribution and under the same licence, and it does not allow incorporation into material released under licences such as MIT or Apache 2.0, nor use in paid courses or other commercial settings. Citing the work does not depend on the licence. Everything is available at \url{https://github.com/pcbrom/homebrewed-slms}, and an archived release with a DOI will accompany a later version of this preprint.

\section*{Acknowledgments}
The authors thank the Laborat\'{o}rio de Processamento e Reconhecimento Inteligente de Sistemas e M\'{e}todos Anal\'{\i}ticos (L-PRISMA) of the Federal Institute of Bras\'{\i}lia, registered in the CNPq Directory of Research Groups, and the Laborat\'{o}rio de Modelo Computacional e Intelig\^{e}ncia para Transporte (TransLab) of the Department of Computer Science, University of Bras\'{\i}lia.

\section*{Use of AI-assisted tools}
AI-assisted tools were used for code development and English language editing. Experimental design, analysis and interpretation are the authors'.

\bibliographystyle{unsrtnat}
\bibliography{references}

\appendix
\section{Proofs}\label{app:proofs}

\paragraph{Proof of \Cref{prop:nmasks}.} Given $t$, the number of masked positions follows $l \sim \mathrm{Bin}(L, t)$ and, given $l$, the masked set is uniform among the sets of size $l$, so the inner expectation of \eqref{eq:nelbo-t} is $\sum_{l=0}^{L} \binom{L}{l} t^l (1-t)^{L-l} F_l$, with $F_0 = 0$. The terms are non-negative, and Tonelli's theorem allows exchanging the sum and the integral,
\begin{equation}
\mathcal{L}(x) = \sum_{l=1}^{L} \binom{L}{l} F_l \int_0^1 t^{l-1} (1-t)^{L-l}\, dt = \sum_{l=1}^{L} \binom{L}{l} B(l, L-l+1)\, F_l .
\end{equation}
Since $B(l, L-l+1) = \Gamma(l)\,\Gamma(L-l+1)/\Gamma(L+1) = (l-1)!\,(L-l)!/L!$, each coefficient is $\frac{L!}{l!\,(L-l)!}\cdot\frac{(l-1)!\,(L-l)!}{L!} = 1/l$, and $\mathcal{L}(x) = \sum_{l} F_l/l = \E_{l \sim U\{1..L\}}[(L/l) F_l]$. The identity was also checked in rational arithmetic for every $L \le 64$ and for every $l$ from 1 to 256 with $L = 256$. $\square$

\paragraph{Proof of \Cref{prop:variance}.} The means agree by \Cref{prop:nmasks}. Under the time form, the marginal law of $l$ is uniform on $\{0, \dots, L\}$, because $\int_0^1 \binom{L}{l} t^l (1-t)^{L-l} dt = 1/(L+1)$, and the posterior of $t$ given $l$ is $\mathrm{Beta}(l+1, L-l+1)$, so $\E[1/t \mid l] = (L+1)/l$ for $l \ge 1$. Given $l$, the set $S$ and hence $f_l$ are independent of $t$. By Jensen's inequality, $\E[(1/t)^2 f_l^2 \mid l, S] \ge (\E[1/t \mid l])^2 f_l^2 = ((L+1)/l)^2 f_l^2$, which is the Rao--Blackwell step. Averaging over $l$ uniform on $\{0,\dots,L\}$, with $f_0 = 0$, gives a second moment of at least $\frac{1}{L+1}\sum_{l=1}^{L} \frac{(L+1)^2}{l^2}\E[f_l^2] = (L+1)\sum_l \E[f_l^2]/l^2$. For the number-of-masks form, $\E[(L/l)^2 f_l^2] = \frac{1}{L}\sum_{l=1}^{L} \frac{L^2}{l^2}\E[f_l^2] = L \sum_l \E[f_l^2]/l^2$. With equal means, the variance of the number-of-masks form is strictly smaller whenever some $\E[f_l^2] > 0$. The same argument applies to each coordinate of the gradient. For the infinite variance of the time form, the event $l = 1$ has probability $L t (1-t)^{L-1}$ given $t$, and on it the squared estimator is $f_1^2/t^2$, so the second moment contains $\int_0^1 L (1-t)^{L-1} \E[f_1^2]/t\, dt$, which diverges at $t = 0$ whenever $\E[f_1^2] > 0$. $\square$

\section{Checks and their state}\label{app:gates}

\begin{table}[!htbp]
\centering
\caption{Checks of the five design decisions and their state. ``Not in this version'' marks checks that depend on people or machines outside the scope of this version.}
\label{tab:gates}
\scriptsize
\begin{tabularx}{\linewidth}{@{}l>{\raggedright\arraybackslash}X>{\raggedright\arraybackslash}p{4.4cm}@{}}
\toprule
Check & Criterion & State \\
\midrule
Formulation G0 & Enumeration of all orders, distribution sums to 1 within $10^{-9}$, NELBO bounds the exact value, injected defects fail & Passed. Defects give 0.5906 and 1.2837 \\
Formulation G1 & Length test through the package iterator, difference within $10^{-9}$ & Passed \\
Formulation G2 & Padding fraction measured before fixing $L$ & $L = 256$, 12.2\% padding \\
Tokenizer G0, G2 & Equal merges and ids, vocabulary rule on the real held-out text & Passed, $V = 4096$ (A), $1024$ (B) \\
Tokenizer G3, G4 & Training and encoding times and equality on the Colab CPU & Not in this version \\
Corpus G1, G3 & Segmentation and minimum block (A), pages, volume and largest group (B) & Passed. B: 444 of 444 pages, 4.76\,MB, largest group 5.78\% \\
Corpus G2 & Manual audit of the class rule on 72 summaries, upper Clopper--Pearson bound below 5\% & Not in this version \\
Corpus G5 & Only configuration and adapter change between A and B & Passed \\
Precision G1 & Sentinel at start and at 20\%, both modes & Passed after the amendment of 26 October 2026 \\
Precision G2 & fp16 versus fp32 interval within $\pm \Delta_{\min}/4$ & Passed \\
Precision G3 & T4 against RTX 5090 & Not in this version \\
Precision G4 & Sentinel active to the final checkpoint & Passed with the sentinel in eval mode \\
Budget G3 & Autoregressive gain from 4 to 16 epochs positive, $\Delta_{\min}$ frozen, Monte Carlo error of the bound below $\Delta_{\min}/4$ & $\Delta_{\min} = 0.0907$. Monte Carlo error checked after the final runs, 0.034 per evaluation with $K = 1$, about 0.014 for the mean of 6 seeds \\
Budget G4 & Bootstrap coverage at least 0.93 & Failed (0.918), randomization test \\
Budget G5 & Decisive budget at 256 epochs & $\Delta = -1.026$, $p = 0.031$ \\
Budget G6 & TOST between T4 and RTX 5090 at 16 epochs & Not in this version \\
Budget G7 & Human reading & Not in this version \\
\bottomrule
\end{tabularx}
\end{table}

\section{Design of the blind reading}\label{app:reading}

The design below is preregistered in \texttt{preregistro/m9\_leitura.json}, SHA-256 \texttt{be22ed9a}, and its results are not part of this version. Three raters read, blind to the origin, crops of 60 words taken from a random position of model samples, of human test excerpts and of word-shuffled excerpts, the last two as controls. A frozen rubric assigns the levels N2 (local syntax), N3 (complete grammatical sentence) and N4 (paragraph coherence). Agreement must reach a Krippendorff's $\alpha$ \citep{krippendorff2011alpha} of at least 0.667, and a level counts at a budget only if at least 10 of 12 items reach it. The automatic measure under study is a local language model acting as judge with an anchored rubric for the same levels, the design that \citet{brom2026trace} used, with a panel of judges and repeated runs, to measure traits of model responses, where anchored semantic rubrics are distinguished from embedding-based evaluators. Grading by a large model is also the evaluation of TinyStories \citep{eldan2023tinystories}. The judge is adopted only if it passes thresholds frozen before the measurement, a lower limit of 0.3 for the 95\% interval of the Spearman correlation per item with the blind reading, at least 80\% of the human ceiling measured by leaving one rater out, and an area under the ROC curve of at least 0.95 separating human text from shuffled and repetitive text.

\section{Hyperparameter search}\label{app:search}

\begin{table}[!htbp]
\centering
\caption{Hyperparameter search on corpus A, mean validation bits per byte of 2 seeds. In bold, the configuration chosen for each mode at 256 epochs.}
\label{tab:search}
\small
\begin{tabular}{llcccc}
\toprule
 & & \multicolumn{2}{c}{Autoregressive} & \multicolumn{2}{c}{Diffusion} \\
\cmidrule(lr){3-4}\cmidrule(lr){5-6}
Learning rate & Dropout & 16 epochs & 256 epochs & 16 epochs & 256 epochs \\
\midrule
$3 \times 10^{-4}$ & 0 & 1.953 & 4.853 & 2.750 & 1.843 \\
$3 \times 10^{-4}$ & 0.1 & 1.931 & 3.668 & 2.756 & 1.863 \\
$3 \times 10^{-4}$ & 0.2 & 1.953 & \textbf{2.850} & 2.763 & 1.930 \\
$10^{-3}$ & 0 & 1.939 & 5.602 & 2.752 & 1.873 \\
$10^{-3}$ & 0.1 & 1.851 & 4.271 & 2.747 & \textbf{1.826} \\
$10^{-3}$ & 0.2 & 1.866 & 3.513 & 2.759 & 1.864 \\
$3 \times 10^{-3}$ & 0 & 2.023 & 5.818 & 2.766 & 2.764 \\
$3 \times 10^{-3}$ & 0.1 & 2.003 & 4.549 & 2.767 & 2.745 \\
$3 \times 10^{-3}$ & 0.2 & 2.010 & 3.745 & 2.765 & 2.359 \\
 
\bottomrule
\end{tabular}
\end{table}

\end{document}